\documentclass[letterpaper,journal]{IEEEtran}

\usepackage{amsmath,amsfonts}
\usepackage{algorithmic}
\usepackage{algorithm}
\usepackage{array}
\usepackage[caption=false,font=normalsize,labelfont=sf,textfont=sf]{subfig}
\usepackage{textcomp}
\usepackage{stfloats}
\usepackage{url}
\usepackage{graphicx}
\usepackage{cite}

\usepackage{booktabs}
\usepackage{multirow}
\usepackage{xcolor}
\usepackage{ulem}
\usepackage{enumitem}       

\usepackage[colorlinks=true, linkcolor=blue, citecolor=blue, urlcolor=blue]{hyperref}
\usepackage[switch]{lineno}

\newcommand{\best}[1]{\textbf{#1}}
\newcommand{\second}[1]{\uline{#1}}

\title{ContextDrag: Precise Drag-Based Image Editing via Context-Preserving
Token Injection and Position-Aligned Attention}

\author{
Huiguo He\textsuperscript{1},
Pengyu Yan\textsuperscript{1},
Ziqi Yi\textsuperscript{1,2},
Weizhi Zhong\textsuperscript{3,2},
Zheng Liu\textsuperscript{2},
Yejun Tang\textsuperscript{2},\\
Huan Yang\textsuperscript{2$\ast$},
Guanbin Li\textsuperscript{4},
Lianwen Jin\textsuperscript{1$\dagger$}
\thanks{$\dagger$ Corresponding author: Lianwen Jin.}
\thanks{$\ast$ Project lead: Huan Yang.}
\thanks{\textsuperscript{1}Huiguo He, Pengyu Yan, Ziqi Yi, and Lianwen Jin are with the
        School of Electronic and Information Engineering, South China
        University of Technology, Guangzhou, China
        (e-mail: hehuiguo@scut.edu.cn; eelwjin@scut.edu.cn).}
\thanks{\textsuperscript{2}Zheng Liu, Yejun Tang, and Huan Yang are with
        KlingAI Research, Kuaishou Technology, Beijing, China
        (e-mail: hyang@fastmail.com).}
\thanks{\textsuperscript{3}Weizhi Zhong is with The University of Hong Kong,
        Hong Kong, China.}
\thanks{Ziqi Yi and Weizhi Zhong are also with KlingAI Research, Kuaishou Technology,
        Beijing, China.}
\thanks{\textsuperscript{4}Guanbin Li is with Shenzhen Loop Area Institute, Shenzhen,
        China.}
}

\begin{document}

\maketitle

\begin{abstract}
Drag-based image editing enables intuitive visual manipulation through point-based drag operations. Existing methods mainly rely on diffusion inversion or pixel-space warping with inpainting. However, inversion inherently introduces approximation errors that degrade texture fidelity, whereas rigid pixel-space operations discard semantic context and produce unnatural deformations.
To address these issues, we introduce \textbf{ContextDrag}, to our knowledge the first framework that brings drag-based manipulation into the in-context image editing paradigm. By leveraging the in-context capabilities of editing models (e.g., FLUX-Kontext), ContextDrag enables precise drag editing without inversion or fine-tuning.
Specifically, we first propose Context-preserving Token Injection (\textbf{CTI}), which injects VAE-encoded reference features into attention layers at spatially aligned target positions, guided by latent-space correspondences estimated directly from user-specified control points. By operating on clean, directly encoded features rather than noisy inversion outputs, CTI preserves rich texture details and enables precise drag control.
Second, we propose Position-Aligned Attention (\textbf{PAA}) to eliminate interference caused by spatial displacement of reference features. PAA re-encodes positional embeddings of displaced reference tokens to match their target locations, and masks out irrelevant reference features at the target location to prevent them from degrading visual consistency.
Experiments on DragBench-SR and DragBench-DR demonstrate that ContextDrag achieves SOTA editing accuracy and overall quality, and comprehensive ablations validate the effectiveness of each proposed component. Code will be publicly available.
\end{abstract}


\section{Introduction} \label{sec:intro}

\IEEEPARstart{D}{rag-based} image editing enables users to adjust visual content by dragging control points to specify spatial changes. This intuitive mechanism supports precise manipulation of object geometry and layout, and has been widely adopted in content creation, digital media production, and interactive design~\cite{NeuralSceneDesigner, TIP_LayoutEditing, DragGAN, TPAMIImageEditingSurvey1, CycleDiff}. With the emergence of diffusion-based generative models~\cite{SD, CLIDE_diffusion, TPAMIImageEditingSurvey2, TPAMIControlSurvey}, this paradigm has gained significant attention.

Existing drag-based editing methods~\cite{FastDrag, diffeditor, lazydrag, geoDrag} typically rely on inversion~\cite{nulltext} to recover intermediate diffusion features that guide the editing process. However, the inversion process is inherently ill-posed, often producing noisy latent features and degrading fine-grained details~\cite{nulltext, FlowEdit, dataWACV, masactrl, Inversions_degrade}. To mitigate this issue, several approaches~\cite{dragdiffusion, gooddrag, draglora} introduce fine-tuning~\cite{dreambooth} to better preserve the appearance of the reference image. This strategy, however, introduces a trade-off between reconstruction fidelity and editability: stronger preservation of the original appearance weakens the effect of editing instructions~\cite{ReNoise, Tight}. As a result, the recovered features often contain unreliable contextual cues, limiting the model’s ability to faithfully preserve fine-grained textures and semantic structures during editing.
Alternatively, Inpaint4Drag~\cite{inpaint4drag} bypasses inversion and performs drag editing through pixel-level manipulations followed by inpainting. While this strategy avoids the instability of inversion, directly manipulating pixels prevents the model from leveraging the semantic and contextual information of the reference image. As a result, such pixel-space operations often produce unnatural deformations~\cite{lazydrag}.

Recently, DiT-based architectures~\cite{DiT} have emerged as a dominant paradigm in diffusion models. OminiControl~\cite{ominicontrol} shows that using the DiT VAE as an image encoder preserves fine-grained textures without task-specific modifications. This strategy has been validated across personalized generation~\cite{UNO, xverse} and general image editing~\cite{FLUX_Kontext, qwen_edit, LongCat}. Among them, FLUX-Kontext~\cite{FLUX_Kontext} achieves SOTA performance in open-source editing through strong in-context modeling. However, leveraging such capabilities for drag-based editing, which requires precise point-driven spatial manipulation, remains unexplored. Adapting these models to drag-based editing poses two key challenges. First, in-context editing models are primarily designed for high-level, text-guided semantic manipulation and therefore lack the spatial precision required for point-driven control~\cite{dragtext, TDEdit, DragFlow}. Second, displacing reference features during dragging misaligns their positional encodings with the target locations, resulting in pronounced geometric distortions in the generated content~\cite{PEF}.

To address these challenges, we propose \textbf{ContextDrag}, a framework that formulates drag-based editing as token-level spatial manipulation within the in-context generation paradigm. Specifically, we first introduce Context-preserving Token Injection (\textbf{CTI}), which estimates latent-space correspondences from user-specified control points and injects reference tokens at their target locations within the attention layers. This design enables direct point-driven spatial control while preserving the full contextual information carried by reference tokens, leading to precise and high-fidelity editing. To resolve the positional-encoding misalignment caused by feature displacement, we further propose Position-Aligned Attention (\textbf{PAA}). PAA consists of two complementary components. Positional Re-Encoding (\textbf{PRE}) reassigns positional encodings to displaced tokens based on their target coordinates, ensuring correct spatial alignment in the attention computation. Overlap-aware Attention Mask (\textbf{OAM}) suppresses conflicting features at target locations, preventing background interference with the dragged content. 
Together, PRE and OAM maintain semantic consistency and preserve texture fidelity in the edited results.

The main contributions of this paper are as follows:
\begin{itemize}[leftmargin=12pt]
    \item We introduce \textbf{ContextDrag}, a novel framework that brings drag-based manipulation into the in-context image editing paradigm, eliminating the need for inversion or fine-tuning.
    \item We propose Context-preserving Token Injection (\textbf{CTI}) to enable precise point-driven spatial control over VAE-encoded reference features within a text-guided generation pipeline.
    \item We develop Position-Aligned Attention (\textbf{PAA}) to maintain spatial coherence when reference features are displaced during editing.
    \item Extensive experiments on DragBench-SR and DragBench-DR demonstrate that ContextDrag achieves SOTA editing accuracy and overall quality. Comprehensive ablations further validate the effectiveness of each proposed component.
\end{itemize}

\section{Related Work} \label{sec:related_works}

\subsection{Drag-based Image Editing}

Drag-based editing~\cite{DragGAN, ling2024freedrag} enables intuitive spatial manipulation by specifying source-target point pairs. For example, DragGAN~\cite{DragGAN} pioneered this paradigm on GANs~\cite{GAN} through iterative motion supervision and point tracking. FreeDrag~\cite{ling2024freedrag} further improves robustness via adaptive point tracking. However, GAN-based methods are inherently limited by the representational capacity and domain specificity of GAN generators.

To overcome these limitations, numerous methods~\cite{dragdiffusion, hou2024easydrag, gooddrag, draglora, SDE_Drag, diffeditor, FastDrag, mou2023dragondiffusion, lazydrag, TDEdit, DragFlow, geoDrag, luo2024rotationdrag, regiondrag, shin2024instantdrag, DragNeXt} have extended drag editing to diffusion models.
DragDiffusion~\cite{dragdiffusion} first adapts this paradigm to diffusion models by recovering latents via DDIM inversion~\cite{nulltext} and preserving appearance through fine-tuning~\cite{dreambooth, LoRA_Composer}. Subsequent works refine this pipeline from multiple angles, including more robust iterative optimization~\cite{gooddrag, hou2024easydrag, draglora} and non-iterative alternatives via attention manipulation, stochastic diffusion, or single-pass warping~\cite{diffeditor, SDE_Drag, FastDrag}. 
Beyond point-level manipulation, GeoDrag~\cite{geoDrag} incorporates depth information to handle complex geometric deformations, and DragFlow~\cite{DragFlow} leverages multimodal LLMs to infer user intent without requiring explicit mask specification. More recently, LazyDrag~\cite{lazydrag} is the first to build drag editing upon DiT~\cite{DiT} architectures, exploiting their stronger generative capacity. 
Despite this progress, all these methods rely on inversion~\cite{nulltext} to recover intermediate latent features. Since inversion is inherently ill-posed~\cite{nulltext, FlowEdit, Inversions_degrade, TIP_BlindInversion}, the recovered features inevitably degrade texture fidelity, and fine-tuning only partially compensates at the cost of editing responsiveness~\cite{ReNoise, Tight}.

Inpaint4Drag~\cite{inpaint4drag} takes a fundamentally different approach by performing pixel-space warping followed by inpainting, thereby bypassing inversion entirely. While this avoids inversion instability, directly operating in pixel space discards the semantic and contextual information of the reference image, producing unnatural deformations under large displacements~\cite{lazydrag}.

In summary, existing drag-based methods face a fundamental structural dilemma. Inversion-based approaches suffer from approximation errors introduced by lossy inversion reconstruction, leading to degraded texture fidelity. Pixel-space approaches avoid inversion but discard reference semantics, forcing the inpainting model to hallucinate missing content. These limitations cannot be fully resolved through incremental improvements within either paradigm.
To address this issue, we adopt a different strategy that operates directly on VAE-encoded reference features, preserving complete contextual information without inversion reconstruction loss. ContextDrag realizes this strategy by injecting VAE-encoded features into attention layers, while Position-Aligned Attention (PAA) mitigates interference caused by spatial displacement.

\subsection{In-Context Image Editing}

Text-based image editing~\cite{gal2022image, diffusionclip, hertz2022prompt, nulltext, MVP, modadapter, kawar2023imagic, wu2023latent, instructpix2pix, CharacterFactory, Parts2Whole, FLUX_Kontext, qwen_edit, LongCat, TPAMIText2ImageSurvey, TIP_POCE, TIP_LayoutEditing} has been extensively studied. Early approaches modify images through learned token embeddings~\cite{gal2022image}, CLIP-guided fine-tuning~\cite{diffusionclip}, cross-attention manipulation~\cite{hertz2022prompt, nulltext}, or instruction-following training~\cite{instructpix2pix}. However, these methods typically rely on inversion or fine-tuning and thus inherit approximation errors that degrade fine-grained details~\cite{nulltext,dataWACV,Inversions_degrade, TIP_BlindInversion}.

Recent DiT-based editing models~\cite{FLUX_Kontext, qwen_edit, LongCat} adopt a fundamentally different strategy: they encode the reference image directly via the VAE and incorporate it into the joint attention computation alongside text and target-image tokens, achieving strong in-context editing capabilities and high-fidelity texture preservation without inversion or fine-tuning~\cite{ominicontrol, ominicontrol2}. This architecture naturally preserves the rich spatial and textural details of the reference image, thereby retaining information that inversion-based methods typically lose.

However, these models lack point-level spatial controllability, as textual descriptions can only provide coarse semantic guidance and cannot specify precise spatial transformations~\cite{dragtext, TDEdit, DragFlow}. ContextDrag bridges this gap by translating point-level drag specifications into token-level operations within the in-context framework, combining spatial precision with high-fidelity feature preservation.

\section{Methodology} \label{sec:method}

\begin{figure*}[t]
  \centering
  \includegraphics[width=1.0\linewidth]{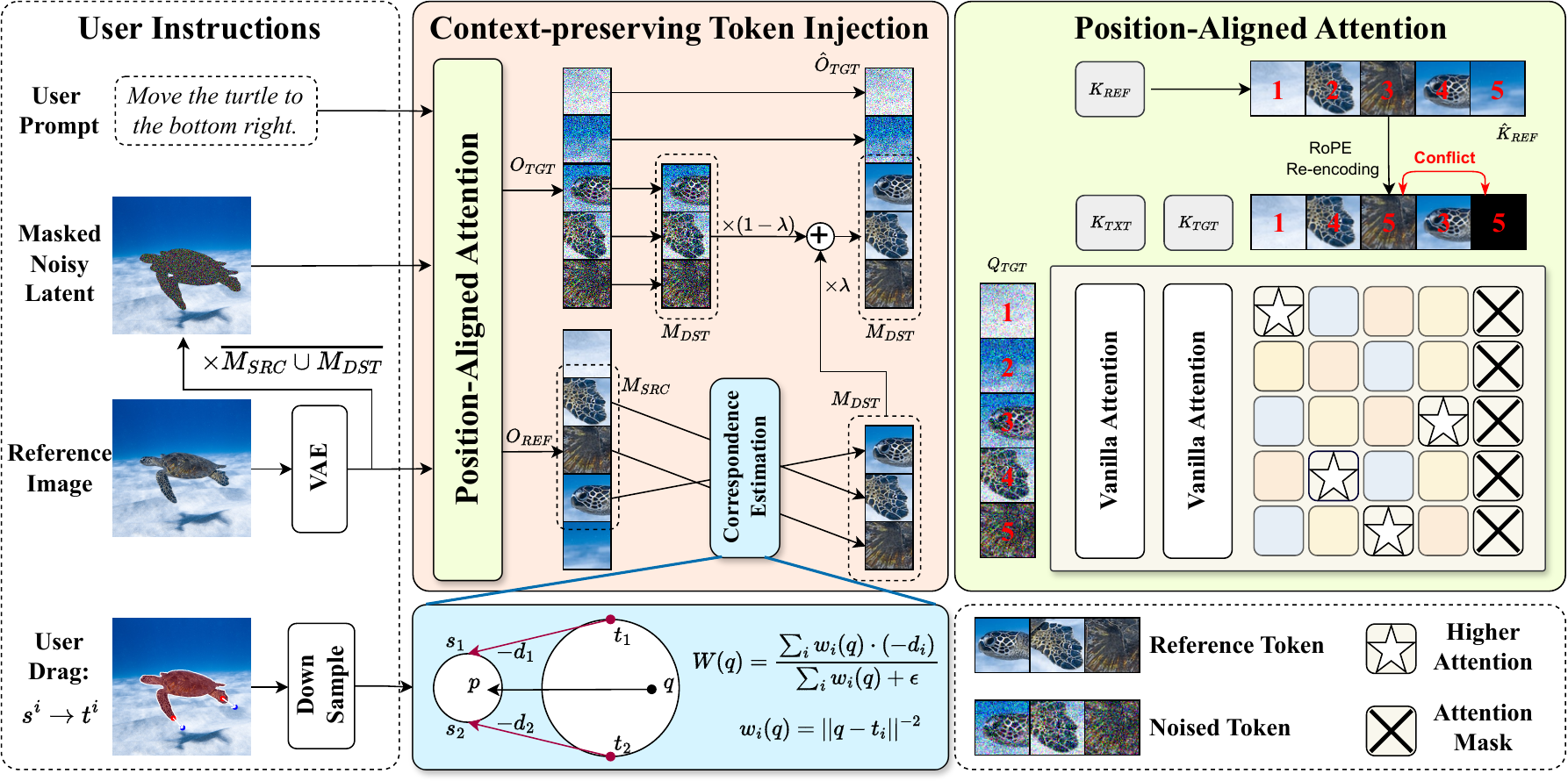} 
  \caption{Illustration of our ContextDrag framework. Context-preserving Token Injection (CTI) directly injects VAE-encoded reference tokens into the correct target positions identified by latent-space correspondence estimation, preserving contextual and fine-grained details. Position-Aligned Attention mitigates interference from reference features through two key mechanisms: (1) it re-encodes the RoPE positional embeddings of reference tokens so that corresponding tokens share corrected positional RoPE and therefore receive higher attention scores; (2) it applies an Overlap-aware Attention Mask to suppress irrelevant signals arising from overlapping regions.}
  \label{fig:framework}
\end{figure*}

\subsection{Preliminaries}

In DiT-based~\cite{qwen_edit, FLUX_Kontext, LongCat} diffusion models, all attention operations are computed jointly over both textual and visual tokens, which differs from that of U-Net-based architectures. A general formulation of the attention layer is given by:
\begin{align}
\label{eq:attn}
    O = Attn(Q,K,V) &= \mathrm{softmax} (QK^{T}) \cdot V.
\end{align}
Here, $Attn()$ denotes the attention function, and $O$, $Q$, $K$, and $V$ correspond to the output, Query, Key, and Value features, respectively. These features are obtained by concatenating the flattened sequences of text tokens and image tokens. $Q$ and $K$ are encoded with Rotary Position Embeddings (RoPE)~\cite{RoPE} to capture spatial information before computing the attention score.

Notably, in editing models (such as FLUX-Kontext~\cite{FLUX_Kontext} and LongCat~\cite{LongCat}), the reference image is directly encoded by the VAE and included in the attention computation. In such a setting, the Query, Key, and Value features are formed by concatenating tokens from the text, the target image, and the reference image: $ O = [O_{\text{\scriptsize \tiny TXT}}, O_{\text{\scriptsize \tiny TGT}}, O_{\text{\scriptsize \tiny REF}}]$, $ Q = [Q_{\text{\scriptsize \tiny TXT}}, Q_{\text{\scriptsize \tiny TGT}}, Q_{\text{\scriptsize \tiny REF}}]$, $K = [K_{\text{\scriptsize \tiny TXT}}, K_{\text{\scriptsize \tiny TGT}}, K_{\text{\scriptsize \tiny REF}}]$, $V = [V_{\text{\scriptsize \tiny TXT}}, V_{\text{\scriptsize \tiny TGT}}, V_{\text{\scriptsize \tiny REF}}]$. 
The subscript `TXT' denotes text features, `TGT' denotes features of the target image, and `REF' denotes features extracted from the reference image.

While these models excel at text-guided editing~\cite{ominicontrol}, they lack the point-level spatial controllability that drag-based editing demands.

\subsection{Problem Formulation}\label{subsec:problem}

Given a reference image $I_{\text{ref}}$, a user-specified source mask $M_{\text{\scriptsize \tiny SRC}}$ indicating the editable region, and $P$ pairs of control points $\{(s_i, t_i)\}_{i=1}^P$ where $s_i$ and $t_i$ denote the source and target positions, the goal of drag-based image editing is to generate an edited image $\hat{I}$ that satisfies two requirements: (1) the content within $M_{\text{\scriptsize \tiny SRC}}$ is spatially transformed according to the drag vectors $d_i = t_i - s_i$, and (2) both the appearance of unedited regions and the texture fidelity of edited regions are preserved.

\subsection{ContextDrag Framework}\label{subsec:framework}

The illustration of our ContextDrag framework is shown in Fig.~\ref{fig:framework}. ContextDrag translates user-provided drag operations into a latent-space warping field that guides precise, context-aware editing. Our approach follows the inpainting paradigm, which is known to best preserve unaltered image regions~\cite{inpaint4drag}. The main challenge lies in (1) estimating accurate and semantically meaningful latent drag vectors, and (2) injecting these deformation signals into the generative process without degrading texture fidelity or contextual consistency.

To this end, our framework operates in two stages.
First, we estimate latent-space spatial correspondences between the source and target regions (Sec.~\ref{subsubsec:warping}), and use these correspondences to inject VAE-encoded reference features into attention layers through Context-preserving Token Injection (CTI), enabling precise and high-fidelity editing (Sec.~\ref{subsubsec:token_inject}).
Second, to address the positional-encoding misalignment introduced by spatial displacement, we develop Position-Aligned Attention (PAA), as detailed in Sec.~\ref{subsec:paa}.

\subsubsection{Latent-Space Correspondence Estimation}\label{subsubsec:warping}

To inject reference features at the correct target positions, CTI requires token-level spatial correspondences that map each position in the post-drag region back to its source in the original image. However, establishing such correspondences in the VAE latent space is non-trivial: unlike pixel space, the latent space is highly compressed (typically $8\times$ spatially), so even small correspondence errors are significantly amplified after decoding. Meanwhile, local smoothness is less critical than in pixel space, because the diffusion model itself provides substantial error correction~\cite{masactrl}. These characteristics dictate that \textit{precise correspondence estimation is more important than enforcing smooth interpolation} in our setting.

Existing correspondence strategies are designed for different operating conditions. Forward warping methods~\cite{FastDrag} compute a displacement for each source point, but this inherently leaves holes in the target region where no source point is mapped~\cite{inpaint4drag}. In our token injection framework, uncovered target tokens lack corresponding reference features entirely, preventing the complete propagation of drag guidance to all target positions. Reverse warping methods~\cite{inpaint4drag} address this by mapping each target position back to a source location, but rely on reference vectors inferred from a forward pass and achieve visual continuity through smooth interpolation, an assumption that does not hold in the compressed latent space.

Based on this analysis, we adopt a two-step strategy. First, we estimate a coarse target region via forward displacement. For each point $p \in M_{\text{\scriptsize \tiny SRC}}$, a provisional drag vector is computed via inverse-distance weighting~\cite{FastDrag}:
\begin{equation}
\begin{aligned}
D(p)= \frac{\sum_i w_i(p) \cdot d_i}{\sum_i w_i(p) + \epsilon},
\qquad w_i(p) = ||p - s_i||^{-2}.
\end{aligned}
\end{equation}
Here, $\epsilon$ is a small constant to prevent division by zero. The displaced point set $\{p+D(p)\}$ induces a cloud of candidate post-drag locations, and taking the convex hull yields a coarse target mask $M^{coarse}_{\text{\scriptsize \tiny DST}}$.

Second, we establish pointwise reverse correspondences within this region. A key observation is that user-specified drag vectors directly encode the intended deformation and are therefore more reliable for correspondence estimation than vectors inferred from a forward pass~\cite{inpaint4drag}, which introduce additional estimation error. For any point $q \in M^{\text{coarse}}_{\text{\scriptsize \tiny DST}}$, its reverse displacement is:
\begin{equation}
\begin{aligned}
W(q)
= \frac{\sum_i w_i(q) \cdot (-d_i)}{\sum_i w_i(q) + \epsilon},
\quad w_i(q) = ||q - t_i||^{-2}.
\end{aligned}
\label{eq:wq}
\end{equation}
We then validate each mapped location $q + W(q)$ by checking whether it falls inside the source region $M_{\text{\scriptsize \tiny SRC}}$. Points whose mapped locations lie outside $M_{\text{\scriptsize \tiny SRC}}$ are discarded, and the remaining valid points form the final target region $M_{\text{\scriptsize \tiny DST}}$. This validation step ensures that every target token has a well-defined source correspondence, which is essential for the subsequent feature injection in CTI.

\subsubsection{Context-Preserving Token Injection}\label{subsubsec:token_inject}

Prior work has demonstrated that injecting features into attention layers through weighted summation is an effective strategy for controllable generation~\cite{masactrl, ConsiStory, dreamstory, CharaConsist}. However, the intermediate features used in these methods have already lost much of the fine-grained visual detail, as they are derived from noisy latent states produced by fine-tuning, inversion, or synchronized denoising.

In contrast, we inject the attention outputs from the reference-image features. These features preserve rich spatial and contextual details because they are derived from the VAE-encoded reference image without noise. We also observe that the editing models (such as FLUX-Kontext) can leverage such image-encoded features to achieve high-fidelity and consistent editing results. 
Therefore, the outputs of the attention layer can be calculated as:
\begin{equation}
\begin{aligned}
\hat{O}_{\text{\tiny TGT}} = O_{\text{\tiny TGT}} &\odot (1-M_{\text{\tiny DST}}) \\
& + \left( \lambda \mathcal{W}_{\Delta}[O_{\text{\tiny REF}}] + (1-\lambda) O_{\text{\tiny TGT}} \right) \odot M_{\text{\tiny DST}},
\end{aligned}
\end{equation}
where $\lambda \in [0,1]$ is a blending coefficient that determines the strength of control. A smaller $\lambda$ retains more of the target’s native attention response, while a larger value emphasizes the reference features. The operator $\mathcal{W}_{\Delta}[\cdot]$ represents a deformation field that spatially warps the reference attention features according to the displacement map. Formally, it samples the reference feature at the source coordinate corresponding to each target position. Specifically, for any target coordinate $q$, 
\begin{equation}
\begin{aligned}
\mathcal{W}_{\Delta}[O_{\text{\tiny REF}}](q)
= O_{\text{\tiny REF}}\big(q + W(q)\big),
\end{aligned}
\end{equation}
where $W(q)$ is the inverse displacement field estimated in the previous subsection. In practice, we implement $\mathcal{W}_{\Delta}$ using direct indexing to sample the corresponding feature values, thereby avoiding interpolation artifacts and preserving the textural fidelity of the reference features.

It is worth emphasizing that our $O_{\text{\tiny REF}}$ are derived from the reference image, which is directly encoded by the VAE and then passed into the DiT model. These features preserve rich texture and details. The editing model can effectively capture this contextual information. As a result, our approach enables accurate drag-based editing and maintains both contextual coherence and texture consistency.

\begin{table*}[t]
\centering
\caption{Quantitative comparison with other SOTA methods on the DragBench-SR and DragBench-DR benchmarks. ContextDrag achieves the best editing accuracy (MD) and the best combined CP$\cdot$PF while matching the highest image fidelity (IF) on both benchmarks. Best performance marked in \best{bold} and second performance marked in \second{underline}.}
\label{tab:SOTA}
\vspace{-6pt}
\resizebox{\textwidth}{!}{%
\begin{tabular}{llcccccccccccc}
\toprule[1.25pt]
\multicolumn{1}{l}{\multirow{2}{*}{Method}} & \multirow{2}{*}{Venue} & \multirow{2}{*}{Finetuning} & \multicolumn{5}{c}{DragBench-SR~\cite{SDE_Drag}} & \multicolumn{5}{c}{DragBench-DR~\cite{dragdiffusion}} \\
\cmidrule(lr){4-8} \cmidrule(lr){9-13}
 & & & MD{$\downarrow$} & IF{$\uparrow$} & CP{$\uparrow$} & PF{$\uparrow$} & CP$\cdot$PF{$\uparrow$}
 & MD{$\downarrow$} & IF{$\uparrow$} & CP{$\uparrow$} & PF{$\uparrow$} & CP$\cdot$PF{$\uparrow$} \\
\hline
SDE-Drag~\cite{SDE_Drag} & ICLR 2024 & $\checkmark$ & 43.70 & \best{0.88} &0.9050  &0.7150  &0.6471
         & 49.74 & \best{0.91} &0.8426  &0.7622  &0.6422 \\
DragDiffusion~\cite{dragdiffusion} & CVPR 2024 & $\checkmark$ & 32.90 & 0.81 &0.9375  & 0.6700  &0.6281
         & 35.38 & 0.88 & 0.9305 & 0.7829  & 0.7285\\
DragLoRA~\cite{draglora} & ICML 2025 & $\checkmark$ & 22.00 & 0.79 & 0.9025 & 0.7675  & \second{0.6927}
         & 26.59 & 0.87 &0.9012  & \second{0.8220} &0.7408 \\
GoodDrag~\cite{gooddrag} & ICLR 2025 & $\checkmark$ & 21.90 & \second{0.82} & 0.9400  &0.7300  & 0.6862
         & 24.37 & 0.87 &0.9293  &0.8085  & \second{0.7513} \\ \hline
DiffEditor~\cite{diffeditor} & CVPR 2024 & $\times$ & 25.77 & 0.81 &0.8150  &0.7725  & 0.6296
           & 25.85 & \second{0.89} & 0.8427 & 0.8195 & 0.6906\\
FastDrag~\cite{FastDrag} & NIPS 2024 & $\times$ & 25.26 & 0.78 &0.9450  &0.7325  & 0.6922
         & 31.63 & 0.86 & 0.9317 &0.7988  &0.7442 \\
Inpaint4Drag~\cite{inpaint4drag} & ICCV 2025 & $\times$ & \second{20.57} & \best{0.88} & 0.8500  & \second{0.8000}  &0.6800
             & \second{22.69} & \best{0.91} & 0.7829  &0.8122  & 0.6359 \\
\textbf{ContextDrag} & \textbf{This Work} & $\times$ & \best{19.07} & \best{0.88} & 0.9225 & \best{0.8200} & \best{0.7565}
 & \best{21.66} & \best{0.91} & 0.9050 & \best{0.8378} & \best{0.7582} \\
\bottomrule[1.25pt]
\end{tabular}
}
\end{table*}

\subsection{Position-Aligned Attention}\label{subsec:paa}

While the correspondence estimation and token injection described above enable precise spatial manipulation in latent space, we observe that the generated results may still exhibit geometric and texture inconsistencies in certain editing scenarios. Specifically, even with accurate correspondence estimation and context-preserving token injection, the results may show distortions or color discontinuities in the manipulated regions.

We attribute this issue to a misalignment in positional encoding between the reference and target regions. As highlighted in prior analyses~\cite{PEF}, minor perturbations in positional embeddings can significantly affect the geometry of synthesized content. In drag editing, when a region is spatially displaced during dragging, the corresponding reference features retain their original positional encodings, leading to interference in attention computation.

To address this problem, we introduce a \textbf{P}osition-\textbf{A}ligned \textbf{A}ttention (\textbf{PAA}) mechanism that enhances spatial coherence during reference-guided editing. PAA comprises two complementary components:
(1) a \textbf{P}ositional \textbf{R}e-\textbf{E}ncoding (\textbf{PRE}) module that recalibrates the positional embeddings of reference features based on the displacement field, and
(2) an \textbf{O}verlap-aware \textbf{A}ttention \textbf{M}ask (\textbf{OAM}) strategy that suppresses interference from irrelevant reference features at target positions.
Together, these techniques maintain geometric consistency in the generation process, resulting in consistent transitions and more faithful texture preservation after drag editing.

\subsubsection{Positional Re-Encoding}

In DiT-based architectures, Rotary Position Embeddings (RoPE)~\cite{RoPE} are applied to $Q$ and $K$ before attention is computed. Let $K^{0}$ denote the key features before applying positional encodings. After positional re-encoding, the modified reference keys $\hat{K}_{\text{\tiny REF}}$ can be expressed as
\begin{equation}
    \hat{K}_{\text{\tiny REF}} = \text{Re-RoPE}( K^{0}_{\text{\tiny REF}}, \,  q, \, q + W(q) ),
\end{equation}
where $\text{Re-RoPE}(K, q, p)$ applies the rotary positional encoding values from position $q$ to the key features $K$ at position $p$. In our formulation, $p = q + W(q)$, where $p$ represents the original coordinate obtained by reverse warping the destination location $q$. Here $W(q)$ is defined in Eq.~\ref{eq:wq}, and $W(q) = 0$ when $q \notin M_{\text{\tiny DST}}$.
The output features of the target image $O_{\text{\tiny TGT}}$ are then computed as
\begin{align}
\label{eq:attn_rope}
    O_{\text{\tiny TGT}} = Attn(Q_{\text{\tiny TGT}},\, [K_{\text{\tiny TXT}},\, K_{\text{\tiny TGT}},\, \hat{K}_{\text{\tiny REF}}], \, V).
\end{align}

This strategy effectively aligns the positional encoding of displaced reference features with their target locations and suppresses artifacts caused by the original misaligned encodings.

\subsubsection{Overlap-aware Attention Mask}

In drag editing, the source region is relocated to a new destination. However, the reference features at this new position retain their original positional encodings. This can cause interference, as shown in the top-right corner of Fig.~\ref{fig:framework}: after applying RoPE, a subject token is encoded with a new position index (e.g., 5), which conflicts with the token originally encoded at that same position (marked in black). To eliminate this interference, we adopt a straightforward solution following prior work~\cite{dreamstory, IRDiffusion, ConsiStory}: we apply a mask to exclude these conflicting features during attention computation. We denote this masking term as $M_{\text{\tiny A}}$, which is given by
\begin{equation}
    M_{\text{\tiny A}} = [\mathbf{0}_{\text{\tiny TXT}}, \mathbf{0}_{\text{\tiny TGT}}, \mathrm{log \left ( 1 - M_{\text{\tiny DST}}(1-M_{\text{\tiny SRC}}) \right )}].
\end{equation}
Here, $\mathbf{0}_{\text{\tiny TXT}}$ and $\mathbf{0}_{\text{\tiny TGT}}$ are zero-valued masks indicating that text tokens and target-image tokens are left unmasked, i.e., they are fully preserved in the attention computation. Therefore, the final target output features are computed as
\begin{equation}
\label{eq:attn_mask}
    O_{\text{\tiny TGT}} = \mathrm{softmax} \left( Q_{\text{\tiny TGT}} [K_{\text{\tiny TXT}},K_{\text{\tiny TGT}},\hat{K}_{\text{\tiny REF}}]^{T} + M_{\text{\tiny A}} \right) V.
\end{equation}

This strategy effectively suppresses interference from irrelevant regions and improves the consistency of the generated image in terms of texture and color.

\section{Experiment} \label{sec:exp}

\begin{figure*}[t]
  \centering
  \includegraphics[width=1.0\linewidth]{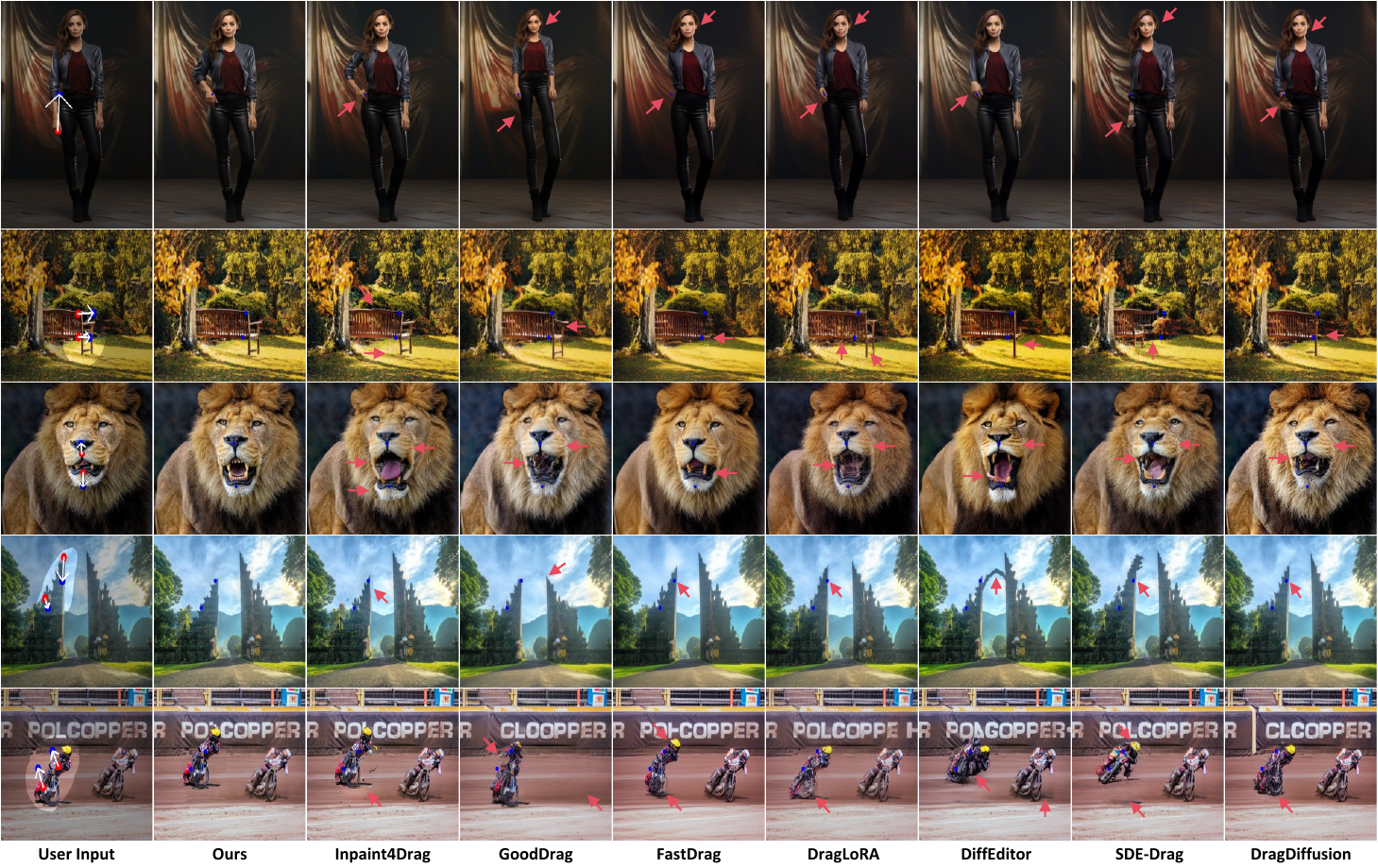}
  \vspace{-20pt}
  \caption{Qualitative comparisons of drag editing between our ContextDrag and other SOTA methods. Our approach achieves the most faithful and consistent edits by effectively leveraging both contextual information and fine-grained texture details. The target points for dragging are marked with blue dots. Additional results are provided in the supplementary material. \textbf{Best viewed with zoom-in.}}
  \label{fig:comparison_visual}
\end{figure*}

\subsection{Implementation Details}

\textbf{Diffusion Model Backbone.}
Since our framework operates on any in-context image editing backbone without modifying CTI or PAA, we evaluate six backbones spanning different model families, scales, and inference regimes. As shown in Tab.~\ref{tab:backbone}, ContextDrag transfers readily across all configurations. FLUX.1-Kontext-dev\footnote{\url{https://huggingface.co/black-forest-labs/FLUX.1-Kontext-dev}} achieves the best overall performance and is adopted as the default backbone for all subsequent experiments unless otherwise specified. Notably, even with the distilled FLUX.2-klein-4B backbone ($\sim$4B parameters, comparable in scale to SDXL), ContextDrag still outperforms most existing methods listed in Tab.~\ref{tab:SOTA}, suggesting that the performance gains stem primarily from our method design rather than merely from a larger backbone.

\begin{table}[t]
\centering
\caption{Backbone generalization on the DragBench-SR benchmark. ContextDrag transfers to different model scales and distilled variants without any modification to CTI or PAA. Best performance marked in \best{bold}.}
\label{tab:backbone}
\vspace{-6pt}
\resizebox{\columnwidth}{!}{%
\scriptsize
\begin{tabular}{lcccccc}
\toprule[1.25pt]
Backbone & Steps & MD{$\downarrow$} & IF{$\uparrow$} & CP{$\uparrow$} & PF{$\uparrow$} & CP$\cdot$PF{$\uparrow$} \\
\hline
FLUX.2-klein-4B~\cite{FLUX2} & 4 & 21.74 & 0.87 & 0.8950 & 0.8075 & 0.7227 \\
FLUX.2-klein-base-4B~\cite{FLUX2} & 30 & 21.99 & 0.88 & 0.9025 & 0.8150 & 0.7355 \\
FLUX.2-klein-9B~\cite{FLUX2} & 4 & 21.22 & 0.87 & 0.9150 & 0.8100 & 0.7412 \\
FLUX.2-klein-base-9B~\cite{FLUX2} & 30 & 20.09 & 0.88 & \best{0.9250} & 0.8150 & 0.7539 \\
LongCat-Image-Edit~\cite{LongCat} & 30 & 20.27 & 0.87 & 0.9225 & 0.8125 & 0.7495  \\
FLUX.1-Kontext-dev~\cite{FLUX_Kontext} & 30 & \best{19.07} & \best{0.88} & 0.9225 & \best{0.8200} & \best{0.7565}  \\
\bottomrule[1.25pt]
\end{tabular}%
}
\end{table}

We set the number of inference steps to 30. The interpolation weight $\lambda$ is initialized at 0.5, held constant for the first 10 steps, then decayed to zero using a cosine schedule over steps 10 to 20, and fixed at zero for the final 10 steps. The RoPE re-encoding and token-injection strategies are applied only to the SingleStreamBlocks (i.e., the latter half of the network), which has been shown to improve consistency~\cite{dreamstory, CharaConsist}.
The user-provided masks are first refined by the SAM~\cite{SAM} model and then processed by the mask pipeline of Inpaint4Drag~\cite{inpaint4drag}. The proposed Overlap-aware Attention Mask (OAM) is applied to all layers to more effectively suppress interference from the reference image. The guidance scale is set to 3.0, and we adopt the default FlowMatchEulerDiscreteScheduler~\cite{SD3} as the sampling scheduler. 
Our ContextDrag is executed in BF16 precision to reduce computational overhead.

\subsection{Benchmark and Evaluation}
\textbf{Benchmark}. We conduct the experiments on two commonly used benchmarks: DragBench-DR~\cite{dragdiffusion} (205 images) and DragBench-SR~\cite{SDE_Drag} (100 images). These benchmarks are well-suited for evaluating drag-editing methods, as they include a variety of drag commands, such as translation, extension, rotation, and deformation. We adopt the Inpaint4Drag~\cite{inpaint4drag} data annotation, in which each sample includes an original image along with a drag diagram indicating the binary mask of the editable region.

\textbf{Metrics}. 
Following previous work~\cite{dragdiffusion, FastDrag, inpaint4drag}, we adopt Mean Distance (MD)~\cite{DragGAN} and Image Fidelity (IF)~\cite{kawar2023imagic} as our primary metrics. MD measures the precision of drag-based editing by computing the Euclidean distance between the target point and the corresponding point in the edited image, with a smaller MD indicating higher accuracy. IF (defined as $1 -$LPIPS) assesses the quality of edited images. A higher IF value reflects greater perceptual and textural similarity to the original image.
To further assess the semantic quality of edited images, we adopt Concept Preservation (CP) and Prompt Following (PF) from DreamBench++~\cite{dreambench_plus}, using GPT-4o as an automatic evaluator. CP evaluates whether the edited image preserves the semantic objects, spatial layout, color, and texture of the original, while PF evaluates whether the correct object is manipulated and reaches the intended target position. For PF, we mark target positions with blue dots in the edited image, as we find that MLLMs exhibit limited sensitivity to subtle spatial displacements without such visual cues. Both metrics are scored from 0 to 4, and we report CP$\cdot$PF as a comprehensive metric for overall performance. The detailed evaluation prompts and case comparisons are provided in the supplementary material.

\textbf{User Study}. To further validate the effectiveness of our method, we conduct a user study employing an A/B testing framework on 305 images from the DragBench-SR and DragBench-DR datasets. Each evaluation is assessed along three dimensions: Prompt Following (PF), Concept Preservation (CP), and Overall. Participants compare two sets generated by different methods and judge which is superior or if they are comparable for each metric. The study collects approximately 1500 votes per comparison, with final results aggregated and presented in percentages.

\subsection{Comparison with SOTA Methods}

We conduct experiments to compare our method with existing SOTA methods, including fine-tuning-based methods (SDE-Drag~\cite{SDE_Drag}, DragDiffusion~\cite{dragdiffusion}, DragLoRA~\cite{draglora}, and GoodDrag~\cite{gooddrag}), and several fine-tuning-free methods (DiffEditor~\cite{diffeditor}, FastDrag~\cite{FastDrag}, Inpaint4Drag~\cite{inpaint4drag}). We exclude LazyDrag~\cite{lazydrag} from the direct comparison because its code is not publicly available; however, we evaluate the inversion-based paradigm that it adopts in the ablation study of Sec.~\ref{subsec:vae_ablation}.

\subsubsection{Quantitative Comparison}

\begin{table}[t]
\centering
\caption{User study results for Concept Preservation (CP), Prompt Following (PF), and Overall. Values are reported as percentages (the \% symbol is omitted). We only show the win and lose results of our model, ties are omitted for clarity. The winning dimension is highlighted in \best{bold}. }
\label{tab:user_study_SOTA}
\small
\resizebox{\columnwidth}{!}{%
\begin{tabular}{lcccccc}
\toprule[1.25pt]
\multirow{2}{*}{Method}
& \multicolumn{2}{c}{CP}
& \multicolumn{2}{c}{PF}
& \multicolumn{2}{c}{Overall} \\
\cmidrule(lr){2-3}
\cmidrule(lr){4-5}
\cmidrule(lr){6-7}
& Win & Lose & Win & Lose & Win & Lose \\
\hline
SDE-Drag~\cite{SDE_Drag} &\best{46.5} &23.5 &\best{58.6} &21.4 &\best{52.3} &13.9 \\
DragDiffusion~\cite{dragdiffusion} &\best{38.6} &26.3 &\best{47.4} &5.3 &\best{50.9} &24.6 \\
DragLoRA~\cite{draglora} &\best{46.8} &27.0 &\best{45.2} &24.4 &\best{45.6} &29.3 \\
GoodDrag~\cite{gooddrag} &\best{35.3} &29.7 &\best{37.0} &21.3 &\best{41.5} &26.0 \\ \hline
DiffEditor~\cite{diffeditor} &\best{53.0} &27.1 &\best{58.2} &32.6 &\best{39.3}&13.4 \\
FastDrag~\cite{FastDrag} &\best{61.1} &22.9 &\best{63.4} &20.9 &\best{61.5} &13.3 \\
Inpaint4Drag~\cite{inpaint4drag} &\best{53.5} &19.1 &\best{33.5} &20.1 &\best{50.0} &17.0 \\
\bottomrule[1.25pt]
\end{tabular}
}
\end{table}

The quantitative comparison with existing SOTA methods is presented in Tab.~\ref{tab:SOTA}. ContextDrag matches the highest IF among all methods while consistently outperforming all baselines in editing accuracy (MD) and the combined CP$\cdot$PF metric, which jointly measures concept preservation (CP) and prompt following (PF).

On DragBench-SR, ContextDrag attains the best MD (19.07, 7.3\% lower than Inpaint4Drag) and the highest PF (0.82).
Although its CP (0.9225) is not the highest, CP tends to be inflated when editing is under-executed~\cite{dragtext, regiondrag, lazydrag}: FastDrag achieves a higher CP (0.9450) simply because its larger MD (25.26) implies smaller displacements that preserve more original content, while Inpaint4Drag matches our IF (0.88) yet suffers a much lower CP (0.8500) due to the loss of semantic context in pixel-space warping. ContextDrag achieves the most accurate edits while preserving both low-level fidelity and high-level semantics, yielding the highest CP$\cdot$PF (0.7565).

On DragBench-DR, ContextDrag again achieves the best MD (21.66) and PF, as well as the highest CP$\cdot$PF (0.7582). Although the margin over GoodDrag in CP$\cdot$PF is modest (0.7582 vs.\ 0.7513), ContextDrag remains superior in the more discriminative metrics, achieving a substantially better MD (21.66 vs.\ 24.37, +11.1\%) and IF (0.91 vs.\ 0.87), all without requiring fine-tuning. We note that existing methods are architecturally tied to U-Net-based diffusion models (SD/SDXL) and cannot be directly applied to DiT-based in-context editing backbones. To isolate the contribution of our method design from backbone differences, we evaluate alternative editing strategies on the same backbone in Sec.~\ref{subsec:vae_ablation}.

The results also reveal a structural difference across paradigms. Inversion-based methods cannot improve editing accuracy without sacrificing image fidelity, as stronger manipulation amplifies reconstruction errors. By operating on inversion-free VAE-encoded features, ContextDrag breaks this trade-off, achieving the best editing accuracy and image fidelity simultaneously.

The quantitative results for different correspondence strategies are shown in Tab.~\ref{tab:ABL_warp}. Our approach consistently outperforms all alternatives across all metrics. Compared with reverse warping using inferred vectors~\cite{inpaint4drag}, our strategy shows a clear improvement in dragging accuracy, achieving an MD of 19.07 (14\% lower than 22.30) and a higher PF of 0.8200 versus 0.8025. Although forward warping~\cite{FastDrag} achieves a reasonable MD, its CP drops substantially to 0.8325 because uncovered target tokens (holes left by the forward mapping) limit the effective editing range, indicating that such approaches struggle to model complex deformation fields in our framework. These results validate our design choice: reverse correspondence with user-specified vectors provides the most accurate spatial alignment for latent-space token injection.

\begin{table}[t]
\centering
\caption{Effect of correspondence strategy on drag editing performance on the DragBench-SR benchmark.
Best performance marked in \best{bold} and second performance marked in \second{underline}.}
\label{tab:ABL_warp}
\resizebox{\columnwidth}{!}{%
\scriptsize
\begin{tabular}{lccccc}
\toprule[1.25pt]
\multirow{2}{*}{Method} & \multicolumn{5}{c}{DragBench-SR~\cite{SDE_Drag}} \\
\cmidrule(lr){2-6}
 & MD{$\downarrow$} & IF{$\uparrow$} & CP{$\uparrow$} & PF{$\uparrow$} & CP$\cdot$PF{$\uparrow$} \\
\hline
Forward + source anchors~\cite{FastDrag} & \second{20.04} & 0.86 & 0.8325 & \second{0.8050} &  0.6702  \\
Reverse + inferred vectors~\cite{inpaint4drag} & 22.30 & \best{0.88} & \second{0.9125} & 0.8025 &  \second{0.7322}  \\
Reverse + user vectors (\textbf{Ours}) & \best{19.07} & \best{0.88} & \best{0.9225} & \best{0.8200} & \best{0.7565}  \\
\bottomrule[1.25pt]
\end{tabular}%
}
\end{table}

\subsubsection{Qualitative Comparison}
In Fig.~\ref{fig:comparison_visual}, we provide a qualitative comparison with existing SOTA methods under several representative editing tasks. Across these examples, our approach achieves more precise manipulations and preserves texture and structural coherence more effectively. In the first example, where the goal is to reposition the woman’s lowered hand to her waist, our method generates a realistic hand structure and maintains consistent character identity. In contrast, competing methods either deform the hand or introduce noticeable identity changes; among them, only Inpaint4Drag~\cite{inpaint4drag} and DiffEditor~\cite{diffeditor} retain the character identity to a limited extent, while all other methods fail to preserve it. Beyond standard edits such as translation and scaling, our method also handles scenarios involving the synthesis of previously unseen content. In the third example, which requires opening the lion’s mouth, our approach reconstructs the missing interior details while preserving the surrounding facial texture. Competing methods struggle to reconcile the dual demands of accurate semantic modification and coherent texture preservation. Similar conclusions can be drawn from Tab.~\ref{tab:user_study_SOTA}.  Overall, these results demonstrate that our approach achieves higher editing fidelity and identity consistency than existing SOTA methods.

\begin{figure}[t]
\centering
  \includegraphics[width=0.8\linewidth]{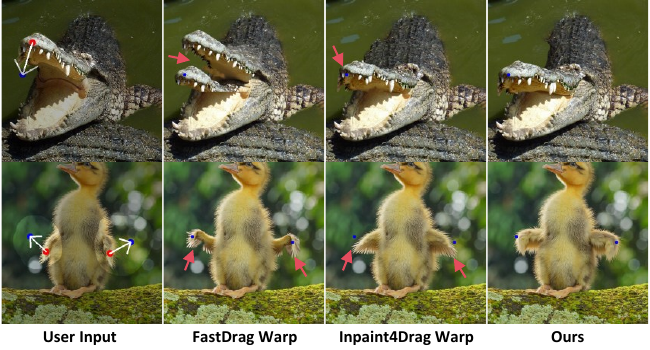}
  \caption{Qualitative comparisons of different spatial correspondence strategies within our framework. Our latent-space correspondence approach achieves superior editing accuracy and semantic consistency. \textbf{Best viewed with zoom-in.}}
  \label{fig:different_warp_visual}
\end{figure}

In Fig.~\ref{fig:different_warp_visual}, we qualitatively compare different correspondence strategies. The results demonstrate that our approach achieves superior editing accuracy and semantic consistency. While forward warping~\cite{FastDrag} achieves partial alignment with the target motion, it often introduces ghosting artifacts or creates structurally implausible results. For example, in the first case, forward warping successfully opens the crocodile’s mouth to the target position but generates duplicated mouth features. Conversely, reverse warping with inferred vectors~\cite{inpaint4drag} maintains structural integrity but sacrifices positional accuracy: in the crocodile example, it fails to open the mouth wide enough to reach the target. These observations confirm that our correspondence strategy, which uses reverse mapping with user-specified vectors in the latent space, outperforms alternatives in both structural reliability and target-position accuracy. Additional correspondence comparisons are provided in the supplementary material.

\subsection{Ablation Studies}

\begin{figure}[t]
\centering
  \includegraphics[width=1.0\linewidth]{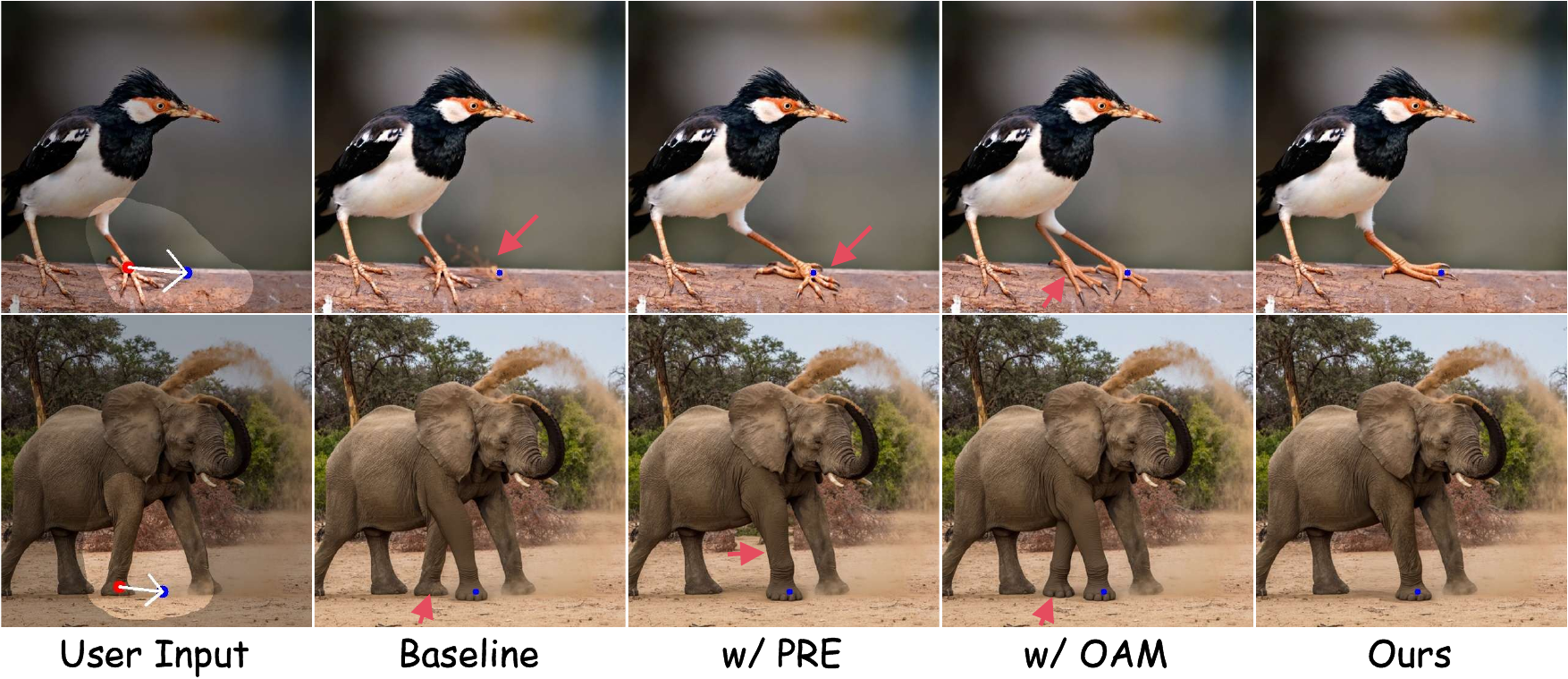}
  \caption{Qualitative comparisons of ablation study. Our full model accurately moves the object to the target destination while perfectly preserving its original appearance. \textbf{Best viewed with zoom-in.}}
  \label{fig:ablation_study_visual}
\end{figure}

\begin{table}[t]
\centering
\caption{Quantitative results of the ablation study on the DragBench-SR benchmark. PRE denotes Positional Re-Encoding, and OAM denotes Overlap-aware Attention Mask.
Best performance marked in \best{bold} and second performance marked in \second{underline}.}
\vspace{-6pt}
\label{tab:ABL}
\resizebox{\columnwidth}{!}{%
\small
\begin{tabular}{lccccc}
\toprule[1.25pt]
\multirow{2}{*}{Method} & \multicolumn{5}{c}{DragBench-SR~\cite{SDE_Drag}} \\
\cmidrule(lr){2-6}
 & MD{$\downarrow$} & IF{$\uparrow$} & CP{$\uparrow$} & PF{$\uparrow$} & CP$\cdot$PF{$\uparrow$} \\
\hline
Baseline & 26.84 & \second{0.87} & 0.8825 & 0.7775 & 0.6861   \\
w/ \, PRE & \best{18.77} & 0.84 & 0.8750 & \second{0.8125} & 0.7109   \\
w/ \, OAM & 22.83 & \best{0.88} & \second{0.9150} & 0.7950 &  \second{0.7274}  \\
w/ \, PRE + OAM (\textbf{Ours}) & \second{19.07} & \best{0.88} & \best{0.9225} & \best{0.8200} & \best{0.7565}  \\
\bottomrule[1.25pt]
\end{tabular}%
}
\end{table}

We conduct comprehensive ablation studies on DragBench-SR~\cite{SDE_Drag} to validate each design choice of ContextDrag. Specifically, we investigate three aspects: (1) the effectiveness of Positional Re-Encoding (PRE) and Overlap-aware Attention Mask (OAM), the two components of our Position-Aligned Attention module (Sec.~\ref{subsec:abl_pre_oam}); (2) the advantage of directly using VAE-encoded reference features over inversion-based alternatives (Sec.~\ref{subsec:vae_ablation}); and (3) the sensitivity to the interpolation weight $\lambda$ in Context-preserving Token Injection (Sec.~\ref{subsec:lambda}). For each study, we present both quantitative and qualitative analyses.

\subsubsection{Effectiveness of PRE and OAM}\label{subsec:abl_pre_oam}

To isolate the contribution of each component in Position-Aligned Attention, we progressively add PRE and OAM to the baseline (CTI only) and evaluate the resulting performance.

\textbf{Quantitative Analysis.}
The results are presented in Tab.~\ref{tab:ABL}. Incorporating PRE alone reduces MD to 18.77, the lowest among all configurations, because it provides strong positional guidance without the masking constraint imposed by OAM. However, both IF and CP drop markedly, indicating poor appearance preservation. This degradation stems from the absence of an attention mask: unintended reference features at the same spatial locations interfere with the generated content after dragging, resulting in inconsistencies in color and texture. Introducing only OAM substantially improves IF and CP by eliminating such interference, yet the MD (22.83) remains suboptimal because the model tends to regenerate the original object in place rather than relocating it when PRE is absent. Combining PRE and OAM yields the best overall performance. The addition of OAM raises MD only slightly (from 18.77 to 19.07) while bringing substantial gains in appearance preservation (CP from 0.8750 to 0.9225, IF from 0.84 to 0.88), resulting in the highest CP$\cdot$PF (0.7565 vs.\ 0.7109 for PRE alone). This validates that the two modules are complementary and both indispensable.

\textbf{Qualitative Analysis.}
As illustrated in Fig.~\ref{fig:ablation_study_visual}, the baseline model struggles with drag editing, producing blurry artifacts and failing to relocate the object or maintain its structure. When equipped with only PRE, the model better adheres to the spatial instruction but compromises visual fidelity, resulting in degraded color and texture. Conversely, OAM alone preserves the object's appearance by shielding it from conflicting reference features, but without PRE's positional guidance, it fails to move the object and instead reconstructs it in place. A successful edit is achieved only when PRE and OAM are combined: the full model accurately relocates the object while preserving its appearance with high fidelity. This synergy confirms that both modules are essential for high-quality drag-based editing.

\begin{table}[t]
\centering
\caption{Comparison of editing strategies on the DragBench-SR benchmark using the same-scale FLUX.1 backbone family. ``Inpainting'' uses pixel-space warping followed by inpainting; ``Inversion'' replaces VAE-encoded features with Uni-Inv~\cite{UniEdit}. Best performance marked in \best{bold}.}
\label{tab:vae_ablation}
\vspace{-6pt}
\resizebox{\columnwidth}{!}{%
\small
\begin{tabular}{lccccc}
\toprule[1.25pt]
Editing Strategy & MD{$\downarrow$} & IF{$\uparrow$} & CP{$\uparrow$} & PF{$\uparrow$} & CP$\cdot$PF{$\uparrow$} \\
\hline
Inpaint4Drag~\cite{inpaint4drag} + FLUX-Fill & 21.71 & 0.87 & 0.8575 & 0.8075 & 0.6924 \\
Inversion$^\dagger$ & 20.33 & 0.81 & 0.7850 & 0.8160 & 0.6406 \\
VAE Encoding (\textbf{Ours}) & \best{19.07} & \best{0.88} & \best{0.9225} & \best{0.8200} & \best{0.7565} \\
\bottomrule[1.25pt]
\end{tabular}%
}
\vspace{1pt}
\raggedright\footnotesize{$^\dagger$ This variant adopts the same inversion-injection paradigm as LazyDrag~\cite{lazydrag}, differing mainly in correspondence estimation.}
\end{table}

\subsubsection{Effect of VAE-Encoded Features}\label{subsec:vae_ablation}

A central design choice of ContextDrag is injecting reference features that are directly encoded by the VAE, rather than recovered through diffusion inversion or pixel-space warping. To validate this choice, we compare our VAE encoding strategy against two representative alternatives on the same backbone family: (1) Inpaint4Drag~\cite{inpaint4drag} combined with FLUX.1-Fill-dev\footnote{\url{https://huggingface.co/black-forest-labs/FLUX.1-Fill-dev}}, which performs pixel-space warping followed by inpainting. Notably, FLUX-Fill is a dedicated inpainting model released by the same developer (Black Forest Labs) as our default backbone FLUX-Kontext, sharing the same FLUX.1 architecture and parameter scale; this ensures that any performance difference reflects the editing strategy rather than backbone capacity. (2) Inversion-based features obtained via Uni-Inv~\cite{UniEdit}, which achieves the highest reconstruction fidelity among existing flow-matching inversion methods.

To ensure a rigorous comparison, we make several adaptations that favor the inversion variant:
(1)~we use 50 inference steps (versus our default 30), as additional steps improve the reconstruction fidelity of Uni-Inv;
(2)~following~\cite{lazydrag}, we fill the vacated source region with Gaussian noise matching the noise level at timestep $T$, which is necessary to prevent the model from regenerating the original content at the source position;
and (3)~we independently tune the injection hyperparameters for the inversion variant ($\lambda = 0.8$, applied during the first 80\% of denoising steps), since the noisier inversion features require stronger and more sustained injection to achieve reasonable editing quality.
These settings are selected to maximize the inversion variant's performance, so that any remaining gap can be attributed to the inherent limitations of the inversion process itself.

\begin{figure}[t]
  \centering
  \includegraphics[width=1.0\linewidth]{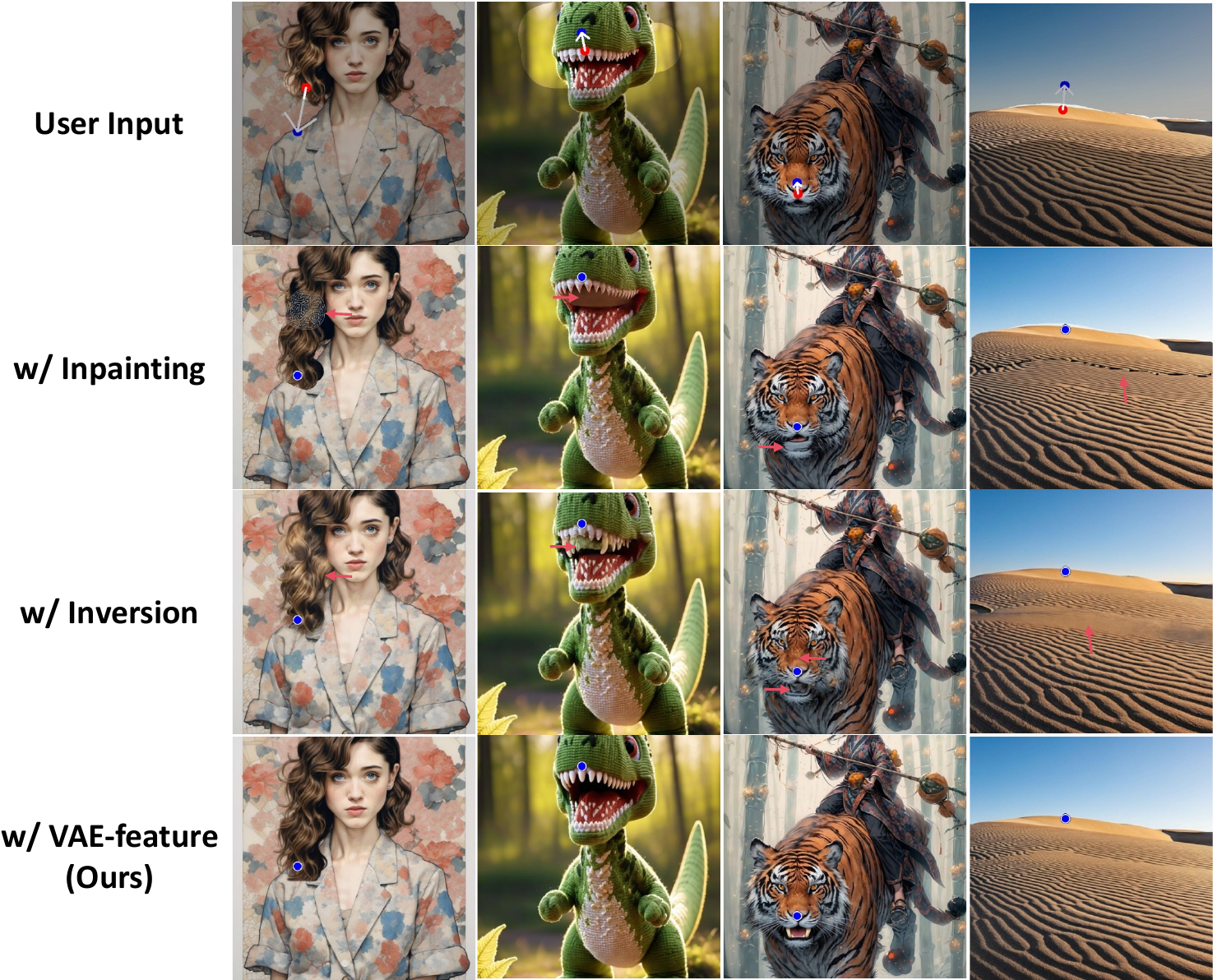}
  \caption{Qualitative comparison of editing strategies. Inpaint4Drag + FLUX-Fill loses semantic context, while inversion (Uni-Inv~\cite{UniEdit}) introduces color shifts and texture degradation. Our VAE-encoded features preserve appearance faithfully. \textbf{Best viewed with zoom-in.}}
  \label{fig:vae_ablation_visual}
\end{figure}

\textbf{Quantitative Analysis.}
Tab.~\ref{tab:vae_ablation} reports the results on DragBench-SR. The Inpaint4Drag + FLUX-Fill baseline, which uses the same backbone family for inpainting after pixel-space warping, achieves reasonable results but still falls behind ContextDrag in all metrics (MD: 21.71 vs.\ 19.07, CP$\cdot$PF: 0.6924 vs.\ 0.7565). This gap confirms that pixel-space operations discard the semantic context that CTI preserves through direct VAE feature injection. Despite the favorable adaptations described above, replacing VAE-encoded features with inversion-based features degrades image fidelity substantially. IF drops from 0.88 to 0.81 and CP from 0.9225 to 0.7850, indicating that inversion fails to retain the fine-grained appearance details preserved by VAE encoding. Editing accuracy also decreases, with MD rising from 19.07 to 20.33 and CP$\cdot$PF falling from 0.7565 to 0.6406. Since this variant shares the inversion-injection paradigm of LazyDrag~\cite{lazydrag}, the degradation reflects a limitation inherent to inversion-based editing. Prior work has suggested that effective drag editing inevitably degrades image fidelity~\cite{dragtext, regiondrag, lazydrag}. However, ContextDrag achieves the best MD while maintaining the highest IF, indicating that this trade-off stems from lossy inversion~\cite{ReNoise,Tight, Inversions_degrade, TIP_BlindInversion} rather than from drag editing itself. Overall, these same-backbone comparisons demonstrate that the performance gains of ContextDrag stem from the CTI+PAA method design rather than from the backbone itself.

\textbf{Qualitative Analysis.}
As shown in Fig.~\ref{fig:vae_ablation_visual}, both alternative strategies exhibit notable limitations. Although Inpaint4Drag + FLUX-Fill employs a dedicated inpainting model, its pixel-space warping discards semantic context and yields less coherent edits. For instance, the opened mouths of the dinosaur and the tiger exhibit undesirable distortions. The inversion-based variant introduces even more pronounced appearance distortions: the girl's hair color is visibly shifted from its original tone, and the sand dune textures are further degraded. In contrast, our method faithfully reproduces both the correct hair color and the intricate sand patterns. These results confirm that CTI and PAA effectively preserve the rich contextual cues that both pixel-space warping and inversion inevitably discard.

\begin{table}[t]
\tiny
\centering
\caption{Ablation study on the DragBench-SR benchmark with different interpolation weight $\lambda$ values. When $\lambda=0.5$, our ContextDrag achieves excellent editing accuracy (MD), the highest Image Fidelity (IF), and the best overall performance (CP$\cdot$PF).
Best performance marked in \best{bold}.
}
\label{tab:lambda_ablation}
\vspace{-6pt}
\resizebox{1.0\columnwidth}{!}{%
\begin{tabular}{lccccc}
\toprule[0.75pt]
$\lambda$ & MD{$\downarrow$} & IF{$\uparrow$} & CP{$\uparrow$} & PF{$\uparrow$} & CP$\cdot$PF{$\uparrow$} \\
\hline
0.1 & 60.99  & 0.90  &  0.9000 & 0.7425  & 0.6683  \\
0.3 & 27.37  & 0.88  & 0.9425  & 0.7600  & 0.7163  \\
0.4 & 20.59  & 0.88  &  0.9400 & 0.7975  & 0.7497  \\
0.5 & 19.07  & 0.88  & 0.9225 & \best{0.8200} & \best{0.7565}  \\
0.6 & 18.81  & 0.87  & 0.9125  & 0.8150  & 0.7436  \\
0.7 & 18.44  & 0.87  & 0.9125  & 0.7925  & 0.7232  \\
0.9 & 19.51  & 0.87  & 0.8650  & 0.7875  & 0.6811  \\
\bottomrule[0.75pt]
\end{tabular}%
}
\end{table}

\subsubsection{Effect of Interpolation Weight}\label{subsec:lambda}

The interpolation weight $\lambda$ in CTI controls the balance between reference-feature injection and the model's native generation. We sweep $\lambda$ from 0.1 to 0.9 to examine its impact.

\textbf{Quantitative Analysis.}
The results are summarized in Tab.~\ref{tab:lambda_ablation}. When $\lambda$ is too small (e.g., 0.1--0.3), the editing signal is insufficient, causing the dragging operation to execute only partially; consequently, both MD and PF metrics degrade.
In contrast, when $\lambda$ is too large (e.g., 0.7--0.9), the injected feature tokens dominate the generation process and hinder the model's ability to correct inconsistencies. This leads to reduced visual coherence, as evidenced by consistently decreasing CP and PF scores despite minor improvements in MD.
Setting $\lambda = 0.5$ achieves the best overall performance: it provides sufficiently strong editing guidance to reduce MD while preserving image fidelity and contextual consistency, yielding the highest CP$\cdot$PF score. We therefore adopt $\lambda = 0.5$ as the default configuration.

\begin{figure}[t]
  \centering
  \includegraphics[width=1.0\columnwidth]{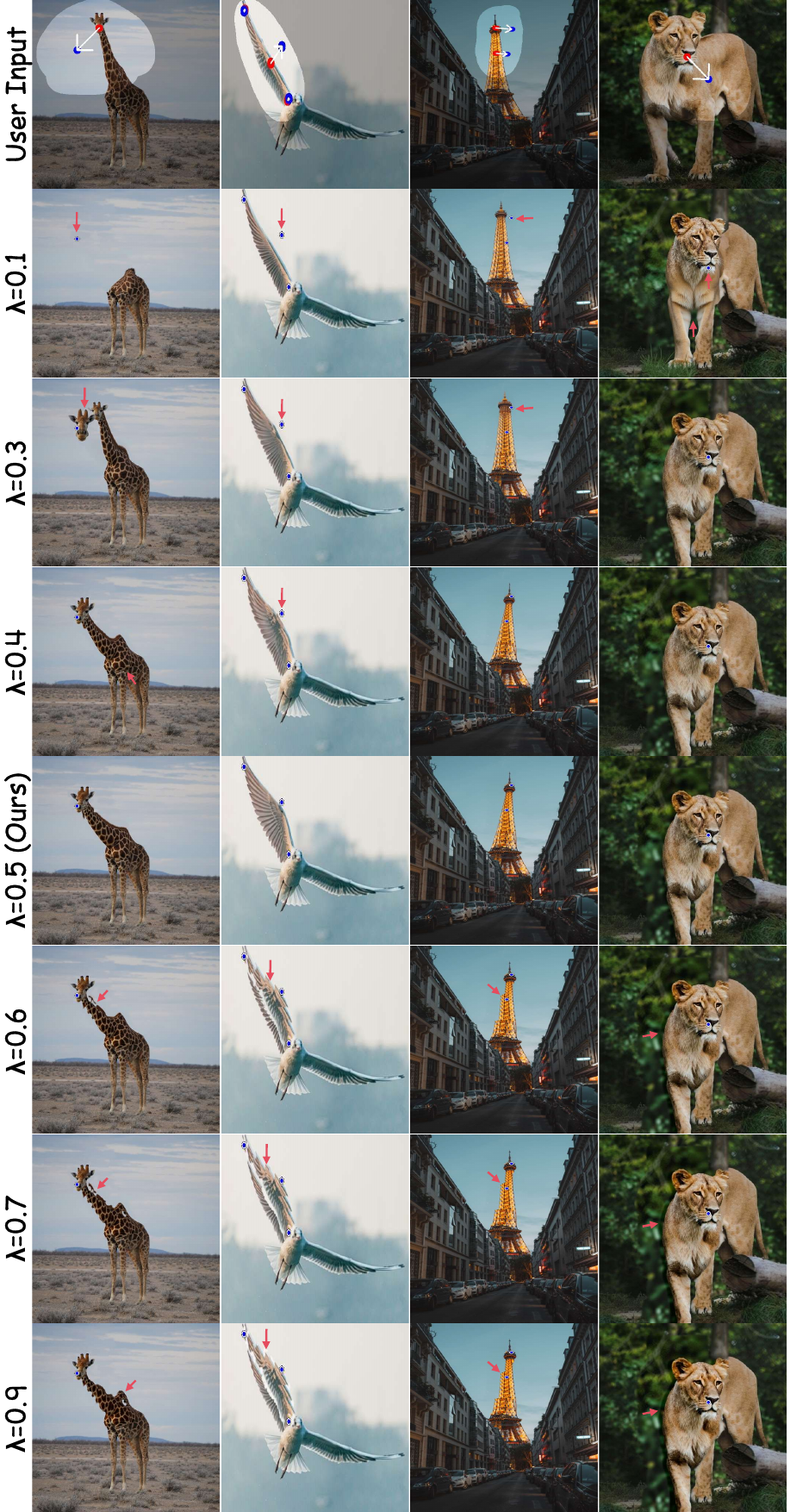}
  \caption{Qualitative comparisons of editing results under different interpolation weights $\lambda$, illustrating the trade-off between editing strength and visual coherence. Our approach achieves the most faithful and consistent edits when $\lambda=0.5$. The target points for dragging are marked with blue dots. 
  \textbf{Best viewed with zoom-in.}}
  \label{fig:interpolation_weight_visual}
\end{figure}

\textbf{Qualitative Analysis.}
Fig.~\ref{fig:interpolation_weight_visual} provides visual comparisons. When $\lambda$ is small ($\leq 0.4$), the editing signal is insufficient to fully execute the dragging operation. As a result, the manipulated content fails to reach the target position. For example, the wing of the bird in the second column and the tilt angle of the Eiffel Tower in the third column remain noticeably under-corrected.
Conversely, when $\lambda$ is large ($\geq 0.6$), the injected features overwhelm the model's corrective capacity, leading to discontinuities and visible artifacts. This ``patchy'' effect is evident in the same examples, where the bird's wing and the Eiffel Tower exhibit abrupt, incoherent transitions.
Overall, $\lambda = 0.5$ provides the best trade-off between editing strength and visual continuity, consistent with the quantitative findings in Tab.~\ref{tab:lambda_ablation}.

\subsection{Efficiency Analysis}\label{subsec:efficiency}

\begin{table}[t]
\centering
\caption{Efficiency comparison. We report Optimization Time (Optim.), Dragging Time (Drag.), and Total Time in seconds. All experiments are conducted on NVIDIA A6000-GPU at a resolution of $1024 \times 1024$.}
\label{tab:Time_SOTA}
\resizebox{\linewidth}{!}{%
\begin{tabular}{llcccc}
\toprule[1.25pt]
\multirow{2}{*}{Method} & \multirow{2}{*}{Venue} & \multirow{2}{*}{FT} & \multicolumn{3}{c}{Time (Seconds) $\downarrow$} \\
\cmidrule(lr){4-6}
 & & & Optim. & Drag. & Total \\
\midrule
SDE-Drag~\cite{SDE_Drag} & ICLR'24 & $\checkmark$ &  866.54 & 52.33 & 918.87 \\
DragDiff.~\cite{dragdiffusion} & CVPR'24 & $\checkmark$ & 245.31 & 141.51 & 386.82 \\
DragLoRA~\cite{draglora} & ICML'25 & $\checkmark$ & 301.90 & 295.97 & 597.87 \\
GoodDrag~\cite{gooddrag} & ICLR'25 & $\checkmark$ & 650.54 & 31.29 & 681.83 \\ \midrule
DiffEditor~\cite{diffeditor} & CVPR'24 & $\times$ & - & 37.45 & 37.45 \\
FastDrag~\cite{FastDrag} & NIPS'24 & $\times$ & - & 8.08 & 8.08 \\
Inpaint4Drag~\cite{inpaint4drag} & ICCV'25 & $\times$ & - & 0.40 & 0.40 \\
ContextDrag$^\ddagger$ & \textbf{This Work} & $\times$ & - & 4.28 & 4.28 \\
ContextDrag & \textbf{This Work} & $\times$ & - & 84.75 & 84.75 \\
\bottomrule[1.25pt]
\end{tabular}%
}
\vspace{1pt}
\raggedright\footnotesize{$^\ddagger$ Using the distilled FLUX.2-klein-4B backbone with 4 inference steps.}
\end{table}

Tab.~\ref{tab:Time_SOTA} compares the efficiency of different methods under 1024$\times$1024 resolution on an A6000 GPU.
As a tuning-free approach, our method eliminates the optimization time required by fine-tuning-based techniques, resulting in faster total execution than all such methods. With the default FLUX.1-Kontext-dev backbone (30 inference steps), ContextDrag takes 84.75s per edit, which is slower than other tuning-free methods due to the larger backbone. To validate that this overhead can be effectively reduced, we apply ContextDrag to the distilled FLUX.2-klein-4B backbone with only 4 inference steps. This reduces the dragging time to 4.28s, a 20$\times$ speedup over our default backbone, while maintaining competitive editing quality (Tab.~\ref{tab:backbone}). This result demonstrates that ContextDrag's efficiency scales favorably with advances in model distillation.

\subsection{Compatibility with Text-Guided Editing}\label{subsec:text_drag}

Since ContextDrag is built upon in-context editing models with native text-guided capabilities (e.g., FLUX-Kontext~\cite{FLUX_Kontext}), our framework naturally supports joint text and drag editing. Our core modules, CTI and PAA, only operate on drag-specified regions to directly inject VAE-encoded reference features and mitigate misalignment artifacts, without interfering with the model's text-conditioned inference pipeline.

\begin{figure}[t]
  \centering
  \includegraphics[width=1.0\columnwidth]{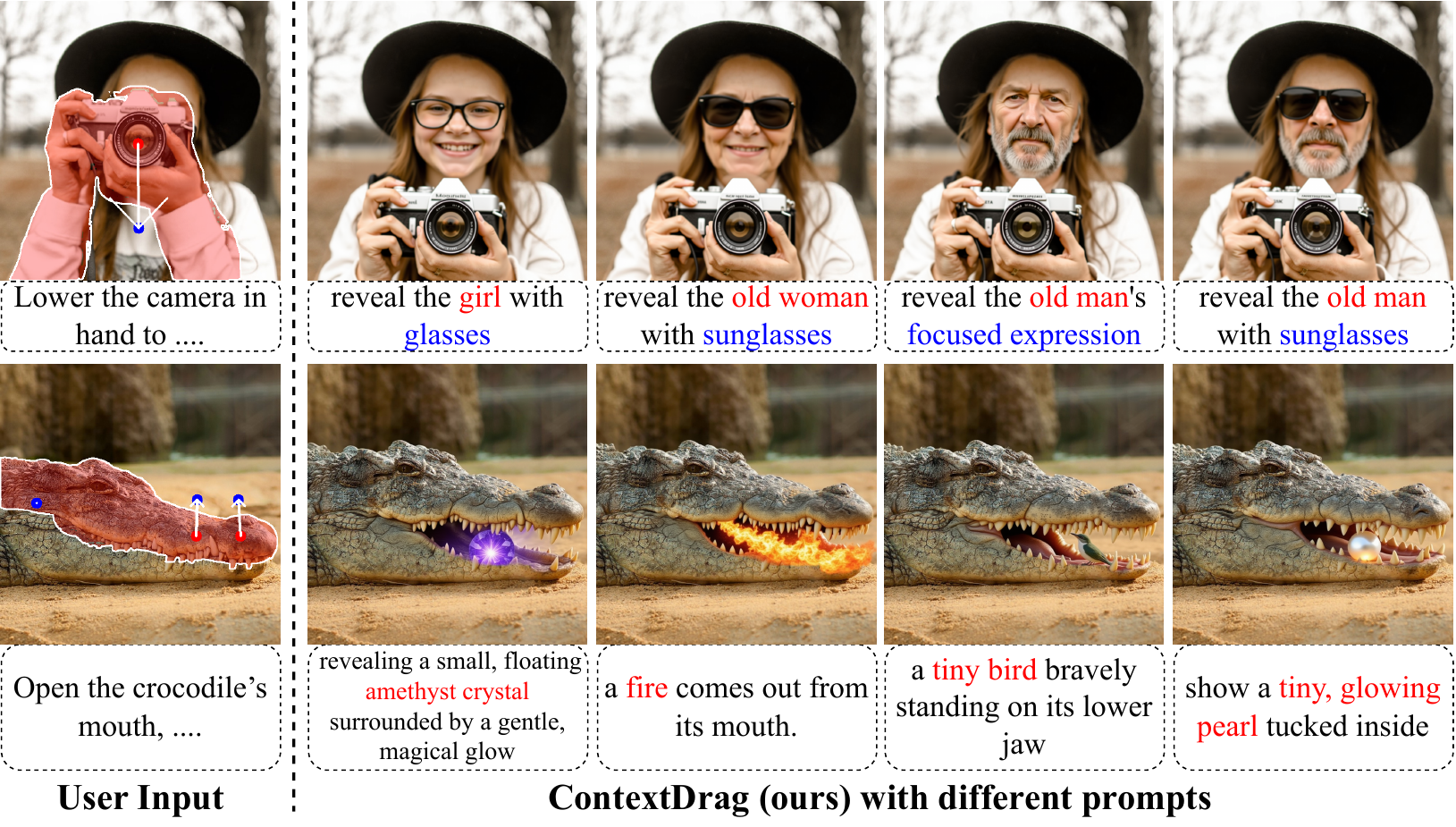}
  \caption{Qualitative results of joint text and drag editing. Given identical drag operations, different text prompts produce diverse semantic outcomes while preserving consistent spatial manipulation. \textbf{Best viewed with zoom-in.}}
  \label{fig:text_drag}
\end{figure}

As shown in Fig.~\ref{fig:text_drag}, given identical drag operations, varying the text prompt produces diverse semantic outcomes. For example, after dragging a hand-held camera downward to reveal the face, different prompts generate distinct identities such as ``girl with glasses'' or ``old woman with sunglasses.'' Similarly, after opening a crocodile's mouth via dragging, different prompts synthesize novel content inside the mouth, such as ``fire'' or ``tiny bird.'' These results show that spatial layout and semantic content can be flexibly controlled via drag and text, respectively, within a single inference pass.

\section{Limitations} \label{sec:limitation}

\begin{figure}[t]
  \centering
  \includegraphics[width=1.0\columnwidth]{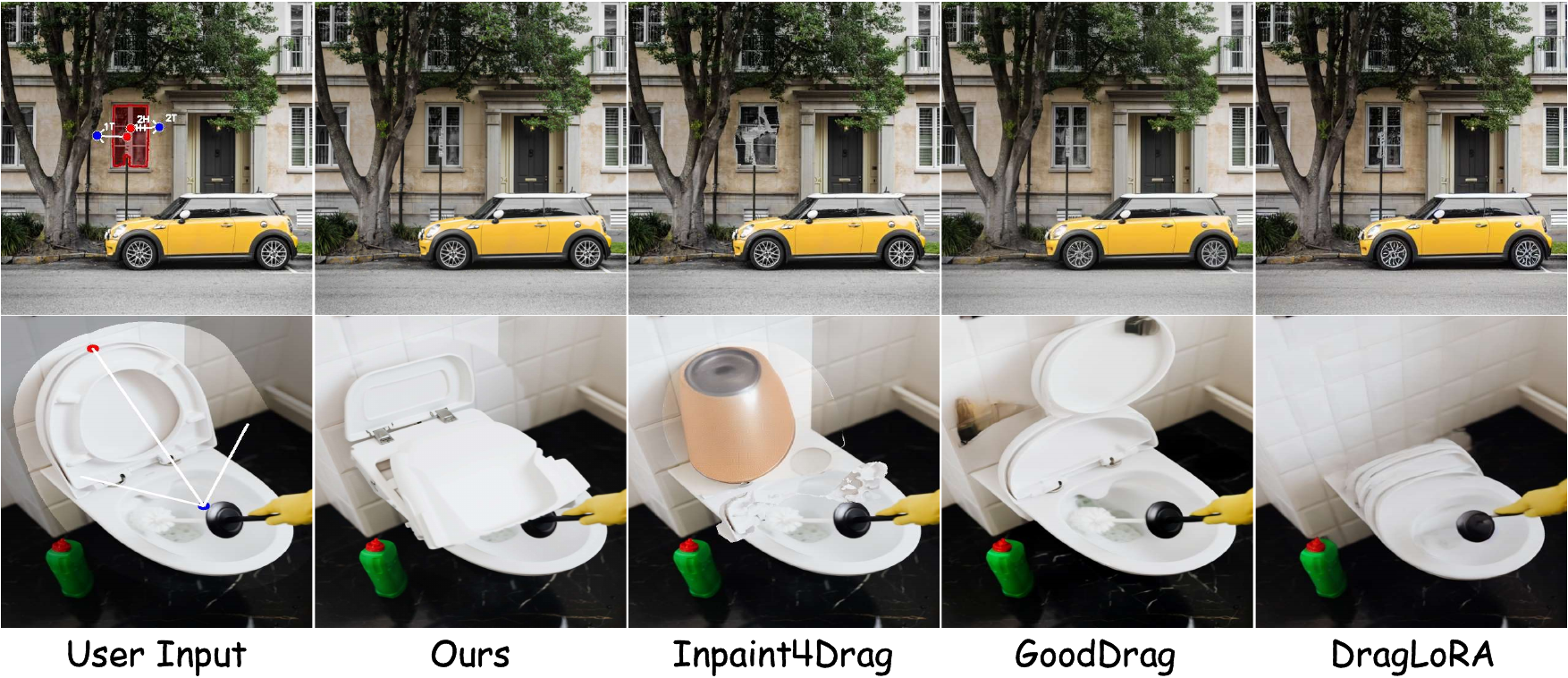}
  \caption{Failure cases of our ContextDrag and competing approaches. Like existing methods, our approach is unable to handle complex deformations, such as flipping. \textbf{Best viewed with zoom-in.}}
  \label{fig:limit}
\end{figure}

Same as most existing methods~\cite{inpaint4drag, gooddrag, draglora}, one limitation of our method is its inability to accommodate highly complex deformations, such as flipping. These complex deformations cannot be effectively represented by limited linear drag vectors from the user, resulting in suboptimal performance in such scenarios. As shown in Fig.~\ref{fig:limit}, none of the methods is able to perform flipping-based drag operations. For instance, in the second row, where the task is to flip the toilet lid downward, all methods fail to accomplish this transformation. We plan to address this limitation in future work.

\section{Conclusion} \label{sec:conclusion}

This work presented \textbf{ContextDrag}, a novel framework that brings drag-based manipulation into the in-context image editing paradigm. By directly incorporating VAE-encoded reference features, ContextDrag eliminates the need for fine-tuning or inversion while preserving high-fidelity visual details. Specifically, we proposed Context-preserving Token Injection (\textbf{CTI}) to enable precise point-driven spatial control by estimating latent-space correspondences and injecting reference features at aligned target positions. To maintain spatial coherence when reference features are displaced, we further developed Position-Aligned Attention (\textbf{PAA}), which re-encodes positional cues and applies overlap-aware masking to suppress interference.
Extensive experiments on DragBench-SR and DragBench-DR demonstrate that ContextDrag achieves SOTA editing accuracy and overall quality, and comprehensive ablations validate the effectiveness of each proposed component. Moreover, ContextDrag readily transfers to other backbones of varying scales with competitive performance, which may inspire future extensions of drag-based manipulation to broader domains such as drag-based video editing, 3D-aware content manipulation, and interactive scene composition.

\section*{Acknowledgment}
This work was supported by the National Natural Science Foundation of China (Grant No.: 62476093) and by Kuaishou Technology.

\bibliographystyle{IEEEtran}
\bibliography{main}

\clearpage
\appendix
\section*{Supplementary Material}

This supplementary material provides additional details and extended experimental results that complement the main paper. Sec.~\ref{supp_sec:cppf} presents the detailed definitions and evaluation prompts of the $CP$ and $PF$ metrics. Sec.~\ref{supp_sec:exp} provides additional visual comparisons with state-of-the-art (SOTA) methods.

\section{Detailed Definitions of CP and PF Metrics} \label{supp_sec:cppf}

This section presents the implementation details of our CP and PF metrics. Both CP and PF are adapted from the metrics proposed in DreamBench++~\cite{dreambench_plus}. These metrics assess not only the consistency between the edited result and the user's instructions (or intent), but also the alignment between the edited and original subjects. These properties are all essential for evaluating drag-editing tasks.

In this section, we introduce CP in Sec.~\ref{supp_subsec:cp} and PF in Sec.~\ref{supp_subsec:pf}. Finally, in Sec.~\ref{supp_subsec:cppf_cases}, we provide several case comparisons to demonstrate that these metrics are well-suited for drag editing.

\subsection{Definition of Concept Preservation (CP) Metric}\label{supp_subsec:cp}

To more comprehensively evaluate drag-based image editing, we adopt an automatic assessment protocol inspired by DreamBench++~\cite{dreambench_plus}, leveraging a multimodal large language model (MLLM) as the evaluator. The goal of the Concept Preservation (CP) metric is to quantify how well the edited image maintains the semantic and visual characteristics of the reference image, independent of the correctness of the dragging operation itself.

\begin{figure*}[t]
  \centering
  \includegraphics[width=1.0\linewidth]{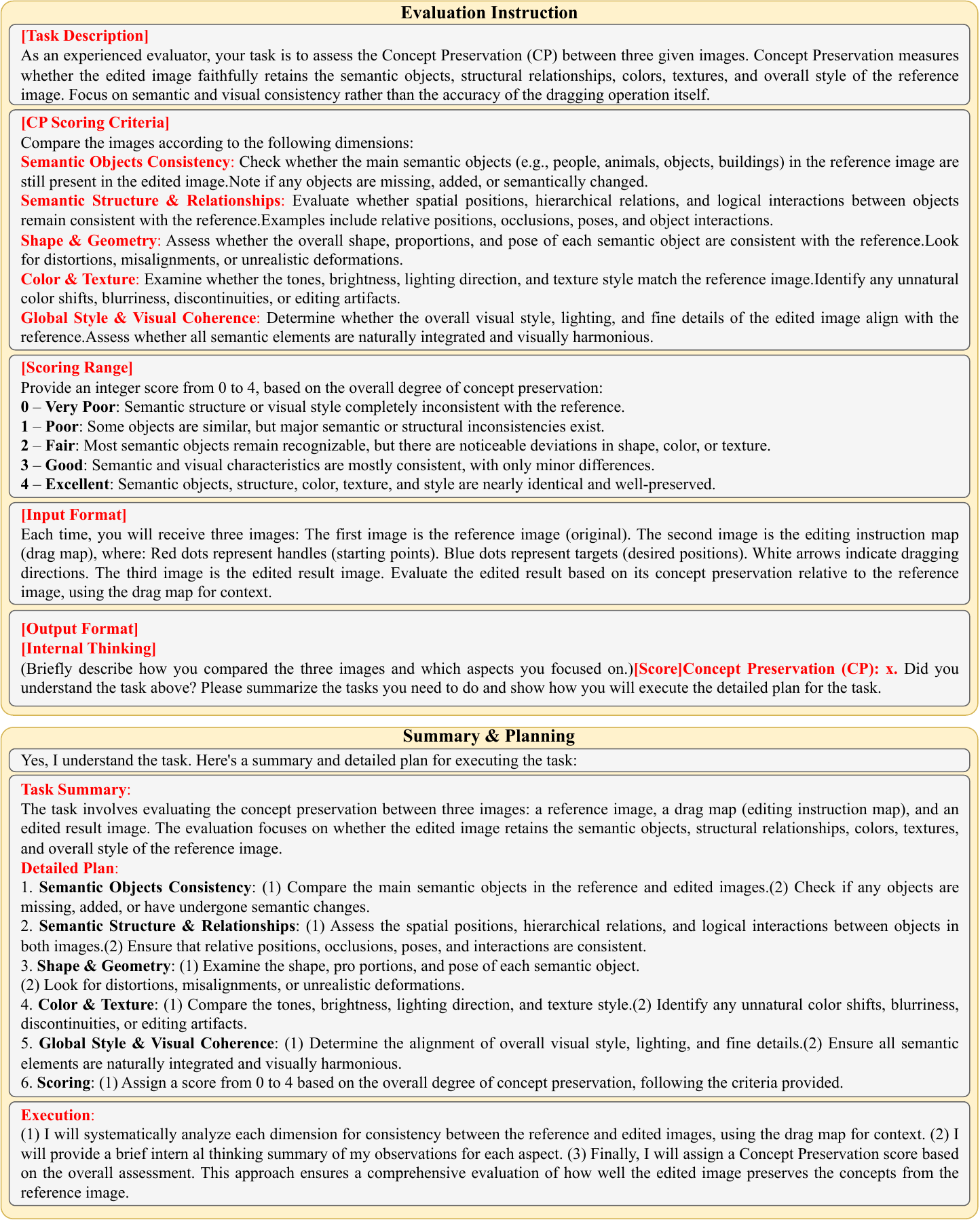}
  \caption{An illustration of the Concept Preservation (CP) Evaluation Instruction and the corresponding Summary \& Planning returned by GPT-4o. The associated code will be publicly available.}
  \label{supp_fig:CP_user}
\end{figure*}

As illustrated in Fig.~\ref{supp_fig:CP_user}, for each test sample we provide the MLLM with three inputs: (1) the \textbf{original image}, (2) the \textbf{drag image} that encodes the editing intent (handles, targets, and dragging directions), and (3) the \textbf{generated image}. The model is instructed to focus on semantic and visual consistency between the original and the generated image, using the drag image only as contextual information about the intended manipulation. Concretely, the MLLM evaluates concept preservation along several dimensions, including:
\begin{enumerate} [leftmargin=16pt]
    \item The presence and consistency of main semantic objects.
    \item The preservation of spatial layouts and object relationships.
    \item The plausibility of shapes and geometry.
    \item The consistency of color and texture.
    \item The global visual style and coherence.
\end{enumerate}
Based on an integrated assessment of these aspects, it outputs an integer CP score ranging from 0--4, where a higher score indicates better preservation of the original concept.

This CP metric enables us to systematically compare different drag-editing methods in terms of how faithfully they maintain the source image's semantics and appearance under user-specified edits, while the detailed prompting and evaluation workflow is deferred to Fig.~\ref{supp_fig:CP_user}.

\subsection{Definition of Prompt Following (PF) Metric}\label{supp_subsec:pf}

To evaluate how faithfully a drag-based editing method follows user-specified intent, we introduce a Prompt Following (PF) metric adapted from DreamBench++~\cite{dreambench_plus}. PF measures whether the edited result correctly interprets the textual prompt and executes the intended manipulation as indicated by the drag map. Unlike perceptual or aesthetic metrics, PF focuses solely on whether the correct object is manipulated, whether it moves in the intended direction, and whether it approaches the designated target position.

\begin{figure*}[t]
  \centering
  \includegraphics[width=1.0\linewidth]{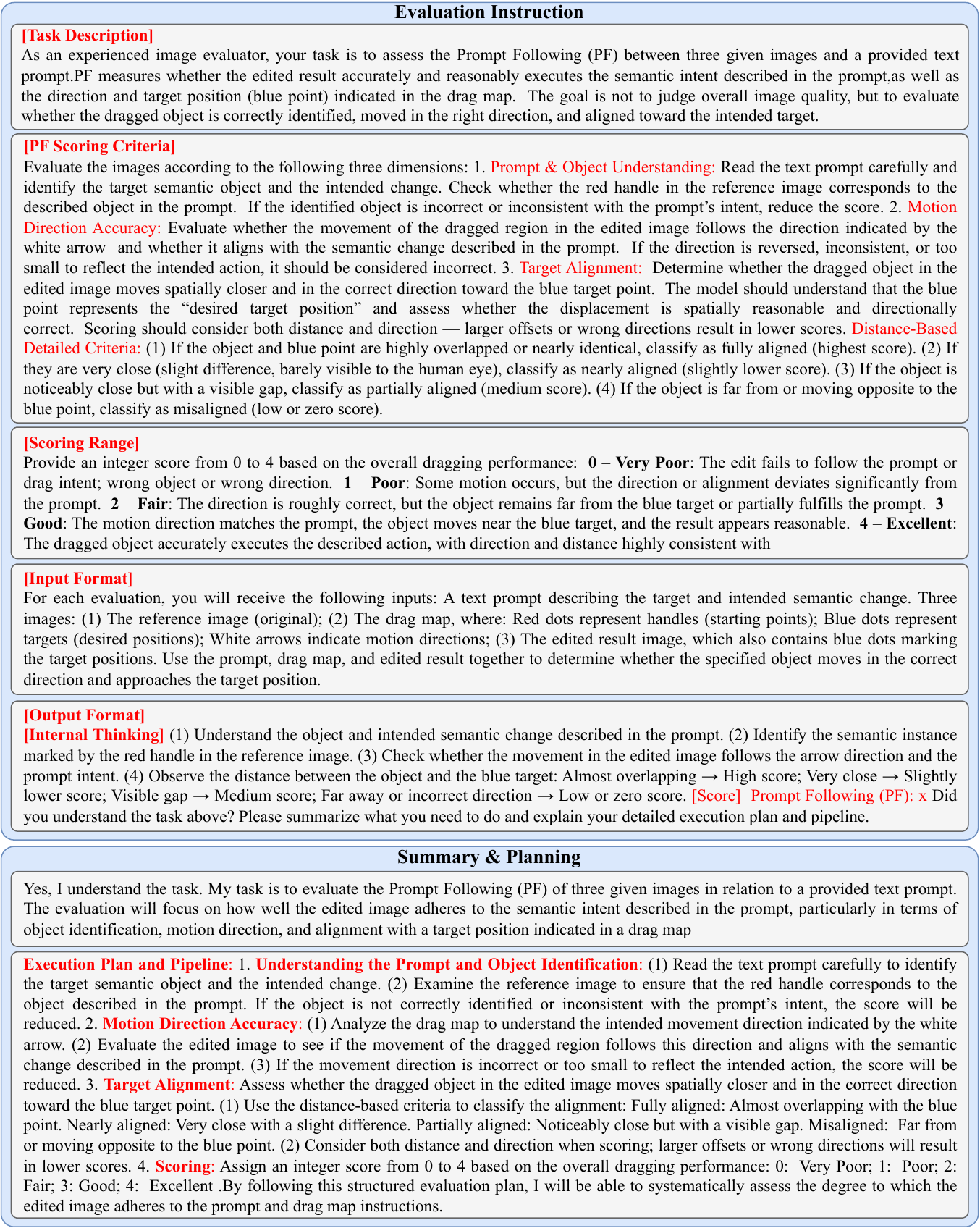}
  \caption{An illustration of the Prompt Following (PF) Evaluation Instruction and the corresponding Summary \& Planning returned by GPT-4o. The associated code will be publicly available.}
  \label{supp_fig:PF_user}
\end{figure*}

As illustrated in Fig.~\ref{supp_fig:PF_user}, the evaluator receives: (1) a \textbf{user prompt} describing the target object and the intended semantic change, (2) the \textbf{original image}, (3) the \textbf{drag image} containing handles (red points), targets (blue points), and directional arrows, and (4) the \textbf{generated image} with blue points marking the drag targets.

The model is instructed to judge how accurately the final image follows the combined intent conveyed by both the text prompt and the drag instructions. Concretely, the MLLM evaluates PF along several key dimensions, including:
\begin{enumerate} [leftmargin=16pt]
\item Whether the correct semantic object specified in the prompt is identified and manipulated.
\item Whether the object's movement direction is consistent with the arrow direction and the semantic intent.
\item Whether the displaced object moves toward and aligns with the specified target point.
\end{enumerate}
Based on an integrated assessment of these aspects, it outputs an integer PF score ranging from 0--4, where a higher score indicates better adherence to the intended manipulation.

We further observe that \textbf{explicitly marking the target positions in the generated image is crucial for reliable PF evaluation}. MLLMs exhibit limited sensitivity to subtle spatial displacements, and without these blue-point annotations, the model often hallucinates incorrect motion directions or misinterprets fine-grained edits. Adding explicit target markers significantly enhances the evaluator's ability to perceive drag-related changes.

This PF metric allows us to systematically quantify how well different drag-editing methods follow user-specified instructions, while the detailed prompting and evaluation workflow is provided in Fig.~\ref{supp_fig:PF_user}.

\subsection{Case Comparison} \label{supp_subsec:cppf_cases}

\begin{figure*}[t]
  \centering
  \includegraphics[width=1.0\linewidth]{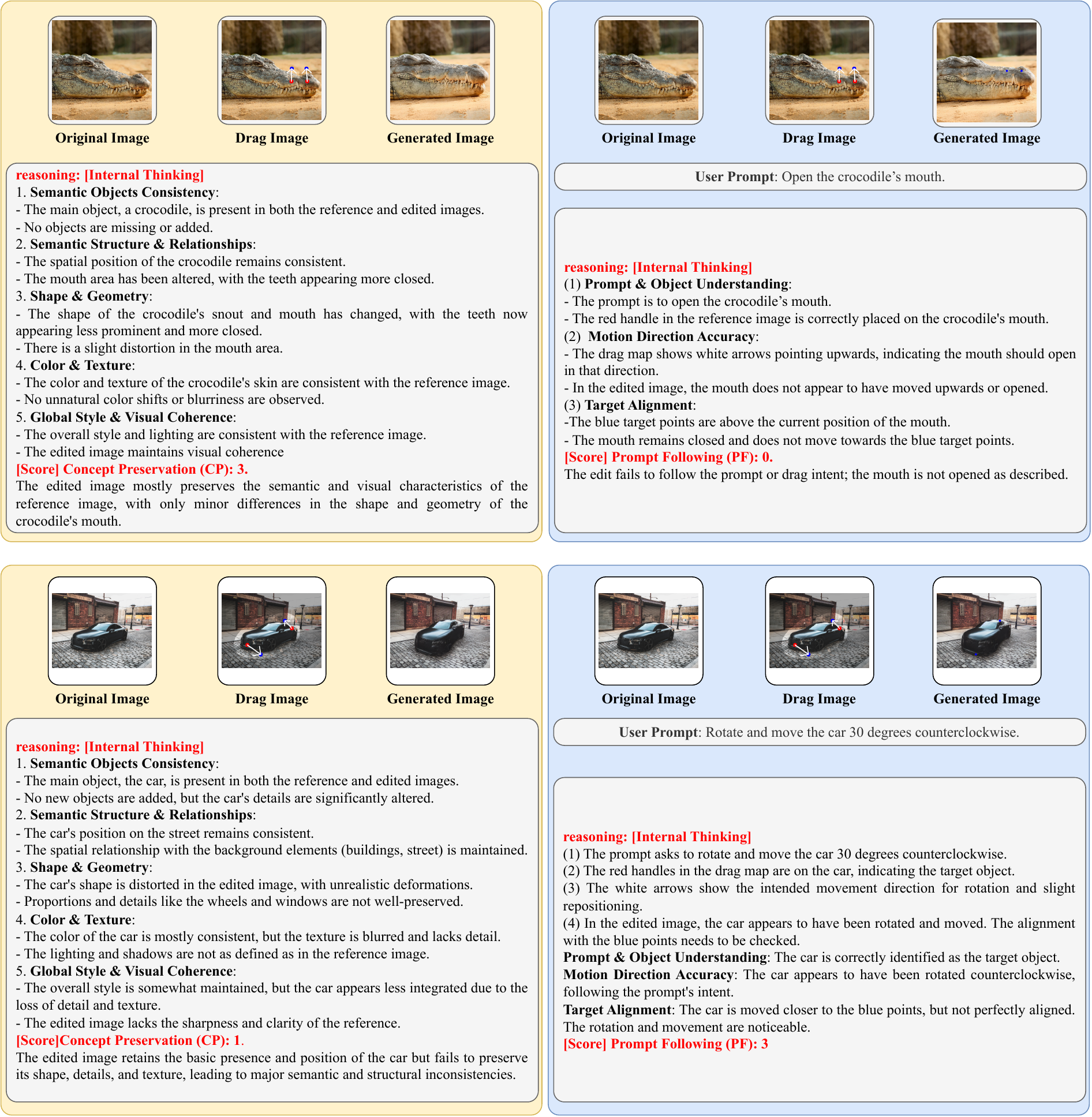}
  \caption{Qualitative examples illustrating both Concept Preservation (CP) and Prompt Following (PF) for a given test case. Each row corresponds to one case, with the left panel showing the Concept Preservation (CP) evaluation and the right panel showing the Prompt Following (PF) evaluation for the same input. The figure highlights how differences in semantic fidelity, visual consistency, motion direction accuracy, and target alignment are reflected in the respective CP and PF scores.}
  \label{supp_fig:caselist1}
\end{figure*}

To further demonstrate the effectiveness of our CP and PF metrics, we present representative qualitative examples in Fig.~\ref{supp_fig:caselist1}. These examples illustrate how the two metrics jointly capture different aspects of drag-based editing quality.

In Fig.~\ref{supp_fig:caselist1} (first row), the crocodile example highlights a case where the editing method fails to follow the user's intended manipulation. Although the scene remains largely unchanged, the crocodile's mouth does not open as specified by the drag instruction, resulting in a PF score of 0. This example shows that PF effectively penalizes cases where the semantic manipulation is not executed, even when the visual appearance remains consistent.

In the same Fig.~\ref{supp_fig:caselist1} (second row), the car rotation example illustrates the complementary role of the CP metric. Here, the object roughly follows the intended motion but undergoes substantial distortion during rotation, resulting in the generated car deviating noticeably from its original appearance. Consequently, the CP score is low (1 point), reflecting poor preservation of the object's shape and visual fidelity despite partial semantic compliance.

Together, these examples validate that CP and PF provide complementary and reliable signals for evaluating drag-based image editing. CP assesses whether the original concept is preserved under manipulation, while PF assesses how well the method adheres to the user-specified intent. Their combined use enables a nuanced and comprehensive evaluation of drag-editing performance.

\section{Additional Experiments} \label{supp_sec:exp}

\subsection{Visual Comparisons with SOTA Methods} \label{supp_sec:sota}
\begin{figure*}[t]
  \centering
  \includegraphics[width=1.0\linewidth]{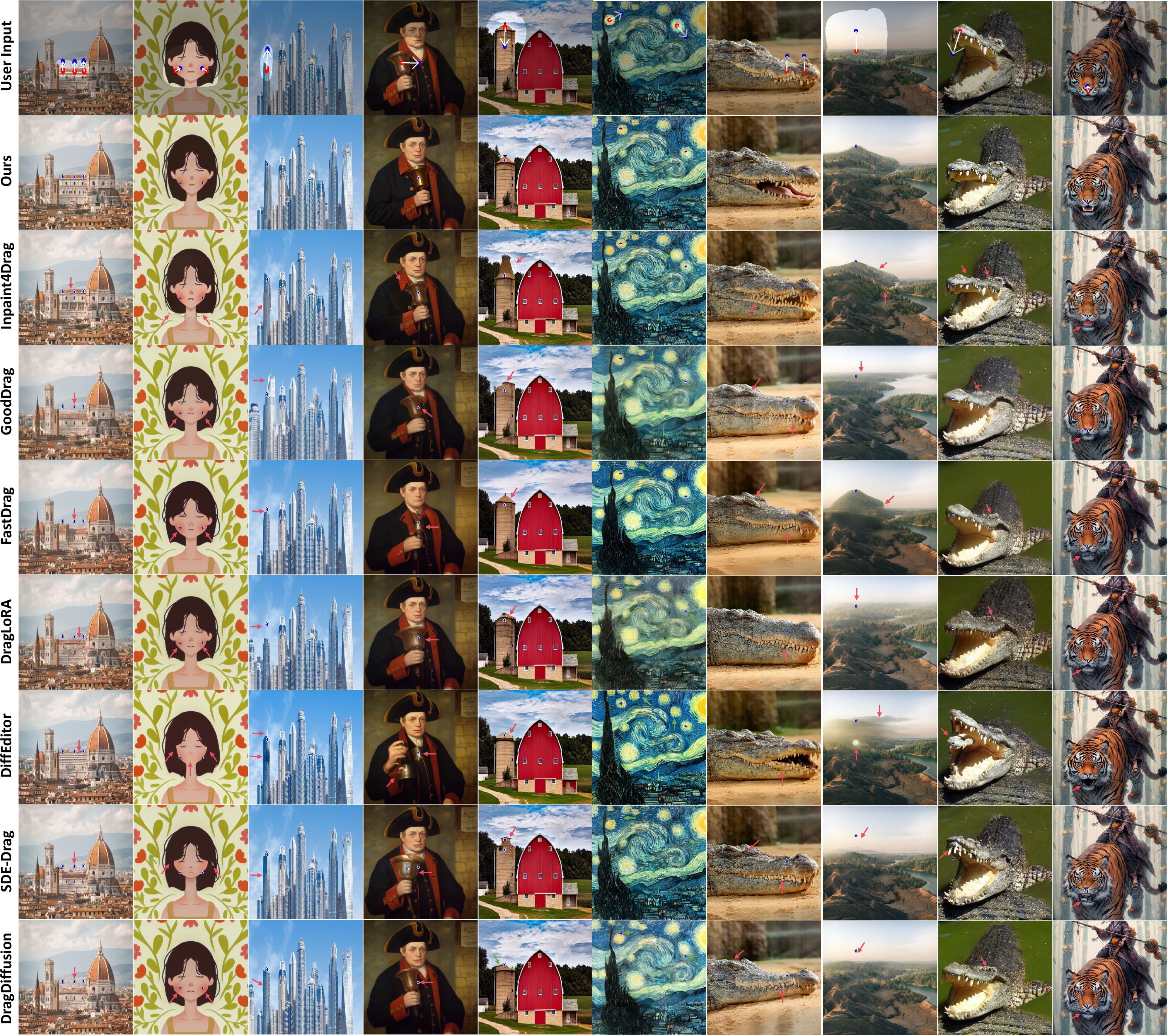}
  \caption{Qualitative comparisons of drag editing between our ContextDrag and other SOTA methods. Our approach achieves the most faithful and consistent edits by effectively leveraging both contextual information and fine-grained texture details. The target points for dragging are marked with blue dots. Regions with inconsistent textures or inaccurate drag editing are highlighted with red arrows. \textbf{Best viewed with zoom-in.}}
  \label{supp_fig:comparison_visual}
\end{figure*}

We provide additional qualitative comparisons with existing SOTA methods in Fig.~\ref{supp_fig:comparison_visual}. Across these examples, our approach achieves more precise manipulations while better preserving both texture details and structural consistency.
For example, in the first case of Fig.~\ref{supp_fig:comparison_visual}, the task is to increase the height of a building. Our method accurately stretches the structure to the target position while preserving its architectural style. Most competing approaches either fail to reach the target height or introduce noticeable stylistic changes; only Inpaint4Drag~\cite{inpaint4drag} and DiffEditor~\cite{diffeditor} achieve a partial stretch but still distort the building's appearance.
Our method also supports edits that require synthesizing previously invisible content. In the last column of Fig.~\ref{supp_fig:comparison_visual}, where the goal is to open a tiger's mouth, our approach realistically reveals the fangs and reconstructs the missing internal details while maintaining coherent surrounding textures. Other methods struggle to balance accurate semantic modification with consistent texture preservation.
Overall, these results demonstrate that our approach achieves higher editing fidelity and identity consistency than existing SOTA methods.

\begin{figure}[t]
  \centering
  \includegraphics[width=1.0\linewidth]{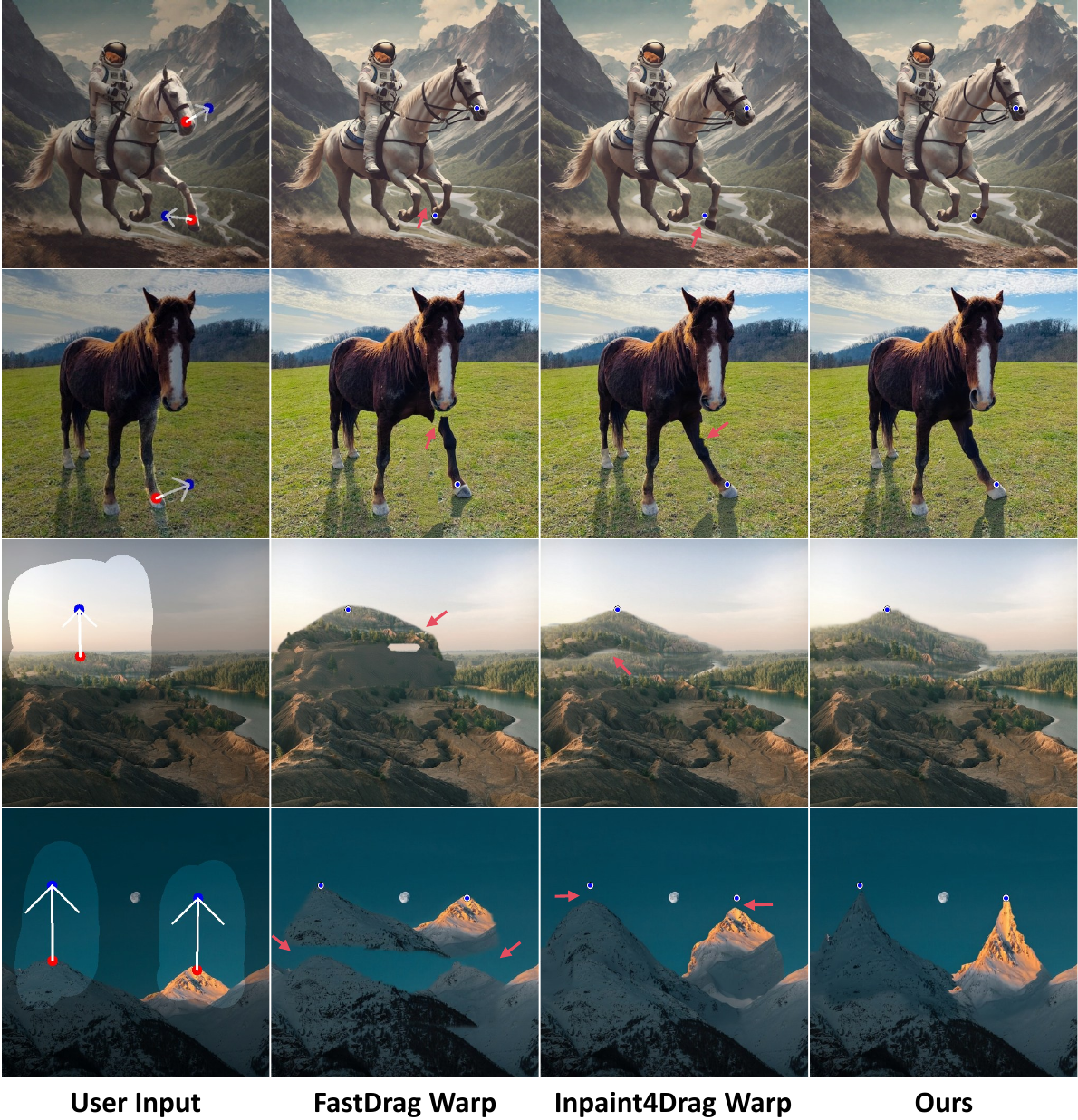}
  \caption{Qualitative comparisons of different spatial correspondence strategies within our framework. Our approach achieves superior editing accuracy and semantic consistency. Regions with inconsistent textures or inaccurate drag editing are highlighted with red arrows. \textbf{Best viewed with zoom-in.}}
  \label{supp_fig:different_warp_visual}
\end{figure}

We also present additional qualitative comparisons of different correspondence strategies in Fig.~\ref{supp_fig:different_warp_visual}. These results also show that our correspondence approach yields superior editing accuracy and semantic consistency.
For example, in the last row of Fig.~\ref{supp_fig:different_warp_visual}, LWF~\cite{FastDrag} fails to handle the complex deformation and incorrectly lifts the entire peak upward, producing an implausible floating structure. In contrast, BWI~\cite{inpaint4drag} preserves structural coherence but lacks positional accuracy, as it does not move the peak sufficiently to reach the target point. Our method successfully resolves both issues, achieving precise alignment with the target while maintaining structural realism.
These findings further confirm that our correspondence strategy outperforms alternatives in both structural reliability and target-position accuracy.

\end{document}